\documentclass[10pt]{article} %
\usepackage[preprint]{tmlr}

\usepackage{amsmath,amsfonts,bm}

\def\eqref#1{equation~\ref{#1}}

\def\1{\bm{1}}

\DeclareMathAlphabet{\mathsfit}{\encodingdefault}{\sfdefault}{m}{sl}
\SetMathAlphabet{\mathsfit}{bold}{\encodingdefault}{\sfdefault}{bx}{n}

\usepackage{hyperref}
\usepackage{url}

\usepackage{microtype}
\usepackage{amssymb}   %
\usepackage{booktabs}
\usepackage{graphicx}
\usepackage{xspace}
\usepackage{todonotes}
\usepackage{multirow}
\usepackage[capitalize]{cleveref}
\usepackage{wrapfig}
\usepackage{lineno}
\usepackage[dvipsnames]{xcolor}

\hypersetup{
    colorlinks=true,
    linkcolor=CadetBlue,
    citecolor=Fuchsia,
}

\newcommand{\method}{{\small\textsc{Mixar}}\xspace}
\newcommand{\methodnospace}{{\small\textsc{Mixar}}}
\newcommand{ \pixar }{{\small\textsc{Pixar}}\xspace}
\newcommand{\pixarnospace}{{\small\textsc{Pixar}}}
\newcommand{ \pixel }{{\small\textsc{Pixel}}\xspace}
\newcommand{ \pixelm }{{\small\textsc{Pixel-M4}}\xspace}

\newif\ifshownotes\shownotesfalse
\ifshownotes
  \newcommand{\av}[1]{\textbf{\color{orange}av: #1}}
  \newcommand{\as}[1]{\textbf{\color{red}as: #1}}
  \newcommand{\chen}[1]{\textbf{\color{blue}chen: #1}}
  \newcommand{\yt}[1]{\textbf{\color{green}yt: #1}}
\else
  \newcommand{\av}[1]{}
  \newcommand{\as}[1]{}
  \newcommand{\chen}[1]{}
  \newcommand{\yt}[1]{}
\fi

\title{MIXAR: Scaling Autoregressive Pixel-based\\Language Models to Multiple Languages and Scripts}

\author{\name Chen Hu$^{\bigstar}$ \email scych4@outlook.com \\
      \addr School of Informatics, University of Edinburgh
      \AND
      \name Yintao Tai$^{\bigstar}$ \email yintao.tai@ed.ac.uk \\
      \addr School of Informatics, University of Edinburgh
      \AND
      \name Antonio Vergari$^{\blacklozenge}$ \email avergari@ed.ac.uk \\
      \addr School of Informatics, University of Edinburgh
      \AND
      \name Frank Keller$^{\blacklozenge}$ \email keller@inf.ed.ac.uk \\
      \addr School of Informatics, University of Edinburgh
      \AND
      \name Alessandro Suglia$^{\blacklozenge}$ \email asuglia@ed.ac.uk \\
      \addr School of Informatics, University of Edinburgh}

\def\month{MM}  %
\def\year{YYYY} %
\def\openreview{\url{https://openreview.net/forum?id=XXXX}} %

\newcommand{\codeurl}{\url{https://github.com/april-tools/mixar}}
\newcommand{\modelurl}{\url{https://huggingface.co/TYTTYTTYT}}
\newcommand{\renderurl}{\url{https://github.com/TYTTYTTYT/Pixar-Render}}

\begin{document}

\maketitle

\makeatletter\if@accepted
\renewcommand*{\thefootnote}{}
\footnotetext{$\bigstar$ Equal contribution.\quad $\blacklozenge$ Joint supervision, correspondence to: Antonio Vergari, Frank Keller, Alessandro Suglia.}
\renewcommand*{\thefootnote}{\arabic{footnote}}\setcounter{footnote}{0}
\fi\makeatother

\begin{abstract}
Pixel-based language models are gaining momentum as alternatives to traditional token-based approaches, promising to circumvent tokenization challenges; however, existing models remain English-centric and are hard to train as generators.
This paper introduces \method, the first generative multilingual pixel-based language model, pretrained on eight high-resource languages spanning Latin and CJK scripts and scaled up to 0.5B parameters.
Although pretrained on high-resource languages only, \method generalizes to unseen low-resource languages better than a tokenizer-based model pretrained on the same corpus, with its largest gains on the scripts most distant from the pretraining mix, taking a step toward more equitable language technology.
To make free generation work in pixel space, \method replaces the notoriously unstable adversarial training stage of prior pixel-based models with a GRPO-based reinforcement-learning stage that rewards the model on its own free-running generations: it proves more stable and multiplies free-generation accuracy by up to $2.3\times$ on multilingual benchmarks.
Finally, \method performs on par with comparable tokenizer-based models on discriminative benchmarks, improves consistently with scale, and withstands orthographic attacks more gracefully than tokenizer-based models.
Code and models are available at \codeurl.
\end{abstract}

\section{Introduction}

A fundamental component in the design of mainstream large language models (LLMs)  is the \textit{tokenizer}, which creates the fundamental input units, i.e., \textit{tokens} that represent the input of the model \citep{touvron2023llama}. 
These elementary units are generally sub-words \citep{wu2016googlesneuralmachinetranslation}, characters \citep{cui-etal-2020-revisiting}, sentence pieces \citep{kudo2018sentencepiecesimplelanguageindependent}, or bytes \citep{sennrich2016neural}. However, creating and maintaining the vocabularies that are associated with these tokenizers is a time-consuming task and is typically done only for a handful of dominant languages~\citep{kharitonov2022textless}. 

To represent tokens as part of an LLM, an embedding matrix,  whose size grows linearly as the number of tokens in the reference vocabulary, is required. 
For modern sub-word LLMs, representing tokens requires millions of parameters, \textit{and this is further exacerbated when dealing with multiple languages}. 
To see why, consider the steady increase of parameters starting from $\sim$23M parameters for a \href{https://huggingface.co/google-bert/bert-base-uncased}{BERT-like model trained only on English},
raising to $\sim$81M for a \href{https://huggingface.co/google-bert/bert-base-multilingual-uncased}{multilingual BERT model}, and $\sim$620M parameters for Qwen3-8B {\citep{yang2025qwen3}}.
Additionally, fixed tokenizers disproportionately harm low-resource languages by degrading performance~\citep{wan2021fairness} and increasing costs~\citep{ahia2023all}, a ``dual penalty'' that widens the global AI equity gap.

\begin{figure}[!t]
    \centering
    \includegraphics[width=\textwidth]{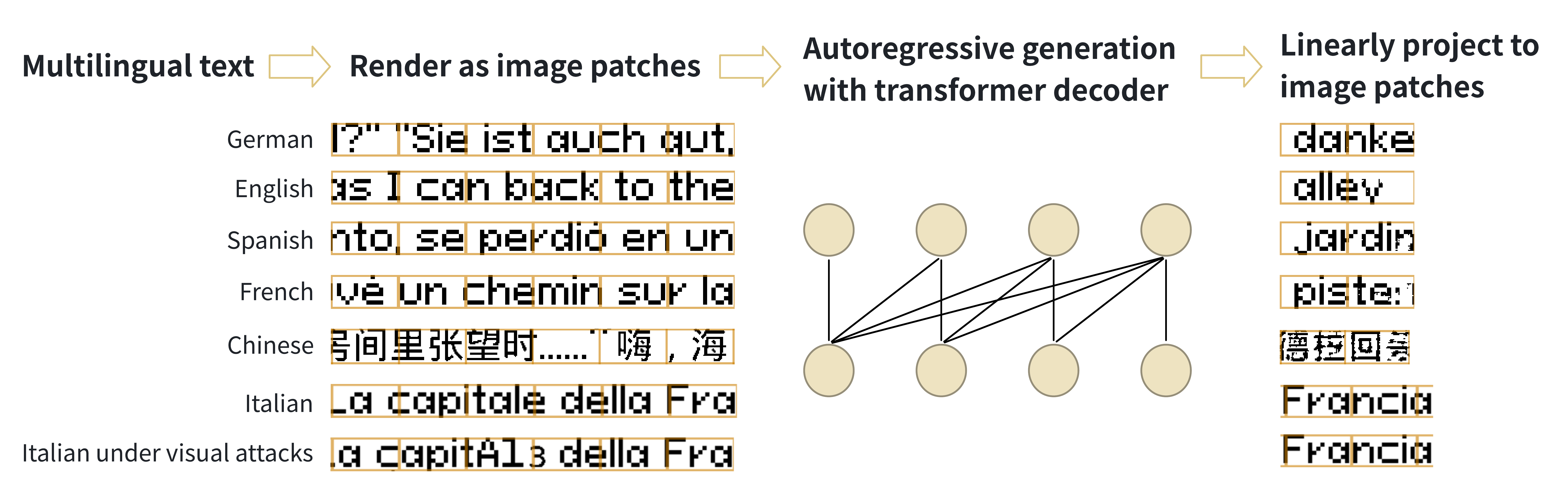}
    \caption{\textbf{\method:} a Transformer-based decoder-only architecture that uses rendered text as input to learn across multiple languages. Encoding language as pixels, enables \method to be robust to visual attacks as well.
    }
    \label{mixar-fig-1}
\end{figure}

Furthermore, token-based LLMs exhibit significant structural vulnerabilities, including the presence of under-trained tokens \citep{land2024fishing}, the phenomenon of unargmaxable tokens \citep{grivas2022low, grivas2024taming} and a general susceptibility to adversarial attacks \citep{shayegani2023survey}.
Recently, \textit{pixel-based models}, i.e., LLMs operating directly (and only) on text rendered as images, such as \pixel \citep{rust2022language} and \pixar \citep{tai2024pixar},
have been proposed as a compelling alternative to token-based models with the promise of ameliorating the aforementioned issues.
In fact, a single pixel-based model should be able to handle multiple languages and scripts without the need to increase its embedding size, as pixels are able to accommodate different writing systems while being more robust to visual changes \citep{rust2022language}.

However, \textit{both \pixel and \pixar were trained only on English}, ignoring the possible benefits pixel-based representations across languages and scripts. This is important if we want to build more inclusive NLP systems that can generalise across languages and cultures~\citep{liu2021visually}.
More recently,  \cite{kesen2025multilingual} introduced \pixelm as a multilingual variant of \pixel, {trained on English (Latin script), Hindi (Devanagari script), Simplified Chinese (Han script), and Ukrainian (Cyrillic script)}.
However, \pixelm, like \pixel, is a discriminative BERT-like model and as such is not able to generate new text (in pixel space).
Although \pixar can generate text autoregressively, one patch at a time, its maximum-likelihood generations suffer from the ``blurry patch'' problem, and making them readable requires an adversarial training stage \citep{goodfellow2014generative} that is notoriously difficult to stabilize.

In this paper, we address these limitations and propose the first \textit{generative and multilingual} pixel-based LLM, \method.
\method is pretrained on multilingual data and replaces the unstable second-stage GAN with a GRPO-style reinforcement-learning stage \citep{mao2026advgrpo} that rewards the model on its own free-running generations, yielding a more stable optimization and stronger generative performance.
Our experiments further show that, although \method is pretrained on high-resource languages only, it generalizes better than tokenizer-based models to unseen low-resource languages with entirely different scripts and cultural contexts, such as Tibetan, Telugu, Kyrgyz and Armenian.
By doing so, we pave the way towards more realistic NLP systems that dispense with tokenizers and instead use other modalities to represent and generate language.

\textbf{Contributions.} We advance the literature on pixel-based language models in the following ways:
\begin{enumerate}
    \item We present \method, a multilingual alternative to \pixar that scales up to 0.5B parameters and higher patch resolutions. \method is pretrained on only eight high-resource languages (\cref{mixar-fig-1}), yet adapts better than tokenizer-based models to unseen low-resource languages from different scripts and cultures, taking a step toward better global AI equity.
    \item We conduct evaluations on both discriminative (e.g.,~natural language inference) as well as generative NLP tasks (e.g.,~language modelling), showing performance competitive with tokenizer-based models like BERT~\citep{devlin2019bert} and superior to the current state-of-the-art pixel-based models.
    \item We propose a new GRPO-based recipe for the second training stage that is more stable than the adversarial pretraining of \pixar and consistently outperforms it on generative tasks.
\end{enumerate}

\section{Beyond English-only and token-based LLMs}

Subword tokenization (e.g., \citealp{sennrich2016neural, schuster2012japanese, kudo2018sentencepiecesimplelanguageindependent}) is the mainstay of most popular LLMs.
However, its effectiveness has been limited to English and a few high-resource languages \citep{touvron2023llama}. 
Additionally, logographic writing systems (e.g., Chinese) become highly challenging for sub-word tokenizers due to their graphic nature. 
In early work, \citet{liu2017learning} used CNNs to capture the visual features of Chinese text at the character level.
On this basis, \citet{sun2018super} used a dataset of rendered Chinese characters to train a text classifier.
Standard symbolic tokenizers often discard the graphical information inherent to Chinese text, as noted by \citet{meng2019glyce}, who used the Tianzige grid format to extract visual features from Chinese characters. Moreover, \citet{dai2017glyph} and ChineseBERT \citep{sun2021chinesebert} integrated character-level visual information into the embedding vectors of BERT-like models. However, these models still commit to a fixed vocabulary and exploit visual information only on the input side. Pixel-based models were designed to tackle exactly these problems.

Using visual information as input, \citet{salesky2021robust} designed a machine translation model, but its output layer still depends on embeddings over a fixed vocabulary.
To overcome this limitation, \pixel \citep{rust2022language} pretrained a masked auto-encoder (MAE) \citep{he2022masked} on a large corpus of rendered text with a masked reconstruction objective.
This model is considered the first pure pixel-based LLM that can handle typical downstream tasks such as extractive question answering and POS tagging.

This study was subsequently expanded to handle multilingual tasks by \citet{salesky2023multilingual}, and further advanced by \citet{kesen2025multilingual}, who introduced PIXEL-M4 to achieve robust cross-script transfer through pretraining on visually diverse languages. Additionally, \citet{tai2024pixar} created \pixar, the first autoregressive pixel-based model, showing that a decoder-only architecture operating purely in pixel space can be competitive with traditional tokenizer-based models.
Pixel-based models have the advantage of representing language directly from rendered text, overcoming the weaknesses of tokenizer-based language models, such as being prone to vocabulary attacks \citep{levi2024vocabulary}.
This benefit was leveraged by \citet{lotz2025overcoming}, who expanded recent LLMs with pixel-based information, demonstrating more robust performance even with logographic writing systems. However, this approach requires ad-hoc expansion techniques, which make training more cumbersome.
Designing and training a scalable and effective multilingual pixel-based generative model is an open challenge.

\section{\textsc{Mixar}}
\label{sec:mixar}

We introduce \method by building upon \pixar, extending its capabilities through four key improvements:
(1)~increasing the patch size to capture visually complex scripts such as Chinese, Japanese, and Korean (CJK) languages (\cref{sec:patch-size});
(2)~scaling the architecture up to 0.5B parameters, demonstrating that pixel-based models are scalable: the larger \method consistently performs better across our benchmarks (\cref{sec:arch});
(3)~curating a diverse training dataset spanning eight languages and multiple scripts (\cref{sec:mutli-pre-train}); and
(4)~conducting a robust multilingual evaluation across various state-of-the-art benchmarks (\cref{sec:exps}).

\subsection{\method architecture}
\label{sec:arch}

\begin{figure}[!t]
  \centering
  \begin{minipage}{.4\textwidth}
  \centering
          \includegraphics[width=5.5cm]{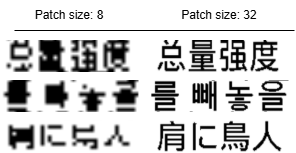}
  \end{minipage}\hfill\begin{minipage}{.55\textwidth}
         \caption{\textbf{Increasing the patch-size in \method helps with non-English scripts} as shown for
         patches of size 8$\times$8 pixels (left) which cannot capture fine-grained details for Chinese (top), Korean (middle) and Japanese (bottom) characters, while 32$\times$32 pixel patches can.}
  \end{minipage}
    \label{3samples4char}
\end{figure}

Similar to \pixar, \method is an autoregressive decoder-only generative LLM that comprises a stack of transformer layers.
Unlike \pixar, we increase the image patch size from 8$\times$8 to 32$\times$32 (see \cref{sec:patch-size}), which is fundamental to capturing non-Latin scripts (see additional details in \cref{3samples4char}) available to the model, but also requires a more expressive decoder.

To this end, we experiment with two different model sizes: a smaller model with 116M parameters containing a stack of 12 Transformer layers (roughly comparable to \pixar's model size), and a larger 477M parameter model featuring 24 Transformer layers.

Following the architectural refinements introduced in LLaMA-2~\citep{touvron2023llama}, we enhance the standard Transformer \citep{vaswani2017attention} by incorporating rotary positional embeddings \citep{su2024roformer}, SwiGLU activation functions \citep{shazeer2020glu}, and pre-normalization using RMSNorm \citep{zhang2019root}. Each Transformer block at layer $i$ outputs hidden states $h^{\text{out}}_i \in \mathbb{R}^{d}$. After the final Transformer layer $N$, a linear layer maps the embedding $h_{N}^{\text{out}}$ back to the pixel space as a vector $\tilde{x}$. In our experiments with binary images, $\tilde{x}$ represents a linearized patch of size $H \times W \times 1$ containing raw logits. To interpret $\tilde{x}$, an element-wise sigmoid function with temperature $T = 1$ squashes the logits into probabilities. %

\subsection{Multilingual pretraining dataset}
\label{sec:mutli-pre-train}
The pretraining datasets used in \citet{tai2024pixar} comprise only Bookcorpus \citep{zhu2015aligning} and English Wikipedia \citep{rust2022language}.
In contrast to these formal and domain-restricted corpora, we utilize the mC4 dataset \citep{raffel2020exploring} to introduce extensive domain diversity through web crawling. 
More crucial to our aim, our corpus encompasses eight high-resource languages including German, English, Spanish, Italian, French, Korean, Chinese, and Japanese.\footnote{Our data sampling and rendering pipeline is part of our released code: \codeurl.} 

We sample documents uniformly from the pooled eight-language mC4 subsets, which follows mC4's original language distribution and avoids oversampling the smaller subsets.
MGPT-2 is pretrained with the same source and the same sampling strategy.

To reduce padding patches, we concatenate consecutive short documents from the same language into a single training sequence until no further document fits, while documents longer than the sequence are split across multiple sequences. A segment-wise (block-diagonal) attention mask restricts every patch to attend only within its own document, so concatenated documents never leak information into each other.
During pretraining, we only apply a length-based filter that discards documents rendering to fewer than two patches.

\subsection{Training regime}
\label{sec:multilingual}
\method is trained in two stages, resulting in two models: \methodnospace$_{\text{stage1}}$, a backbone pretrained with the maximum-likelihood objective of \citet{tai2024pixar}, and \methodnospace$_{\text{stage2}}$, refined with a short GRPO-based free-generation stage.
In the first stage, the model is trained with teacher forcing \citep{williams1989learning}. Conditioned on the preceding gold patches $x_{1:i-1}$, it minimizes the negative log-likelihood of the next $L$ ground-truth patches $x_{i:i+L-1}$.
We reshape each output embedding from the final transformer layer into a single-channel patch. \citet{tai2024pixar} observed that the single-channel prediction outperforms RGB. We then compute the reconstruction loss $\mathcal{L}_{\text{rec}}$ as the pixel-wise binary cross-entropy between the predicted and the ground-truth patches \citep{kingma2013auto, ghosh2019variational}.

This objective has a structural flaw for generation. Every pixel of a patch is predicted independently given the same output embedding. When the next patch is uncertain, the per-pixel optimum is therefore the average of all plausible continuations, not a readable text patch.
As a result, \methodnospace$_{\text{stage1}}$ generates blurry, noisy patches, and the degradation compounds as the model free-runs beyond one block.
We therefore run a short second pretraining stage that directly targets readability.
\citet{tai2024pixar} used a GAN-based second stage. In our experience, this setup is notoriously unstable and hard to tune. In our early explorations, it took 44 runs to find one that did not degrade the stage-1 performance.
We instead refine the model with reinforcement learning based on group-relative policy optimization (GRPO) \citep{grpo}. This training proves both more stable and more effective than the GAN (\cref{app:stage2ablation}).

From a randomly selected word onset, the generator autoregressively samples a group of $G$ candidate continuations of $K$ patch blocks each. Every rollout $y$ is scored by a reward with two parts:
\begin{equation}
r(y) \;=\; d_{\text{blend}}\,\sigma\big(D(y \mid x)\big) \;+\; (1-d_{\text{blend}})\,\big[\lambda_{\text{tile}}\,\text{tile}(y, x) + (1-\lambda_{\text{tile}})\,\text{pix}(y, x)\big],
\label{eq:reward}
\end{equation}
where $D$ is a discriminator, initialized from \methodnospace$_{\text{stage1}}$, that judges the realness of each generated patch given the preceding context $x$, and $\sigma$ is the sigmoid. Let $x^{*}$ be the ground-truth continuation of $x$. Then $\text{tile}(y, x)$ is the fraction of patches of $y$ that exactly match $x^{*}$, and $\text{pix}(y, x)$ is the fraction of matching pixels. $d_{\text{blend}}$ and $\lambda_{\text{tile}}$ weight the two parts.
The discriminator part penalizes the blurry patches that MLE favors. The ground-truth part penalizes realistic-looking but wrong glyphs. It recycles the stage-1 supervision in reward form: the same reference continuation now scores the model's own samples instead of teacher-forced predictions. 

Within each group, rewards are standardized into advantages $A$, and the generator minimizes
\begin{equation}
\mathcal{L}_{\text{stage2}} \;=\; \lambda_{\text{pol}}\,\mathcal{L}_{\text{PPO}}(A) \;+\; \beta_{\text{KL}}\,\text{KL}\big(\pi \,\|\, \pi_{\text{ref}}\big) \;+\; \lambda_{\text{rec}}\,\mathcal{L}_{\text{rec}},
\label{eq:stage2loss}
\end{equation}
where $\mathcal{L}_{\text{PPO}}$ is the clipped PPO policy objective, $\pi$ is the generator, and $\pi_{\text{ref}}$ is the frozen stage-1 model. $\mathcal{L}_{\text{rec}}$ is the teacher-forced reconstruction loss of stage 1, kept as an anchor to the data manifold. 
$\lambda_{\text{pol}}$, $\beta_{\text{KL}}$ and $\lambda_{\text{rec}}$ weight the terms. Together with the reward weights, they were selected in pilot sweeps on validation LAMBADA free generation (search ranges and per-model values in \cref{app:stage2rldetails}).

Finally, the generator and the discriminator are trained in alternating phases rather than jointly, for two reasons.
First, the discriminator is initialized from \methodnospace$_{\text{stage1}}$ and cannot judge realness yet. An initial discriminator-only phase therefore trains the reward model before the first policy update.
Second, the alternation keeps the discriminator competent but not dominant. A dominant discriminator scores every rollout close to zero, and the reward becomes too sparse to rank the rollouts within a group.
Formally, let $a_t$ be an exponential moving average of the discriminator's accuracy with decay $0.95$. Generator updates run while $a_t \ge \tau_{\text{hi}} = 0.75$. Once the improving policy erodes $a_t$ below $\tau_{\text{lo}} = 0.55$, generator updates pause and the discriminator is retrained until $a_t$ recovers above $\tau_{\text{hi}}$. Each phase is bounded to $100$--$400$ steps, which prevents oscillation and keeps the discriminator from over-training. We set these values once in early pilot runs on validation LAMBADA and reuse them unchanged for all three models.
This \emph{waiting} mechanism ensures the generator only receives rewards from a competent judge and avoids the joint-update instabilities typical of adversarial reinforcement learning. 
Using an adversarially trained reward inside GRPO has recently been explored by Adv-GRPO \citep{mao2026advgrpo}. The waiting mechanism is what sets our algorithm apart.

\section{Experiments}
\label{sec:exps}

We aim to answer the following research questions: 
(\textbf{RQ1})~Does \method generalize to unseen low-resource languages better than tokenizer-based models?
(\textbf{RQ2})~Is the GRPO stage-2 training more stable and effective than adversarial pretraining?
(\textbf{RQ3})~Are pixel-based LLMs scalable, i.e., does performance consistently improve as \method grows in capacity?
(\textbf{RQ4})~How robust is \method against orthographic attacks compared to tokenizer-based models?

\subsection{Experimental setting}

\textbf{Datasets.} 
We pretrain on the packed mC4 mix of \cref{sec:mutli-pre-train}, for a total of $138$B patches. This is exactly the patch budget of \pixar, which spends it entirely on English, whereas English accounts for only ${\sim}$17.4B (12.6\%) of our patches. Since sequences are packed, we quantify the data distribution by the number of language-specific patches the model observes (\cref{tab:dataset_stats}). The breakdown also reveals the encoding efficiency of pixels on non-Latin scripts: a single patch encodes 3.67--4.94 GPT-2 tokens for CJK languages, against 0.81--1.20 for Latin ones.

\textbf{Text rendering.}
To mitigate the font rendering issues of previous work~\cite{rust2022language,kesen2025multilingual}, we developed a new rendering tool for \method that produces bit-identical renderings across machines, released as the standalone package \texttt{pixar-render} (\renderurl): instead of relying on the host's font stack, which silently falls back to system default fonts and can change glyphs between environments, we re-initializes the font engine with a fixed, bundled font set and never touch system fonts. Text is drawn with the pixel-style font \href{https://www.dafont.com/pixeloid-sans.font}{Pixeloid Sans} at font size $32$, and three bundled fallback fonts (Go Noto Current, Noto Sans Symbols~2, Noto Emoji) extend coverage to CJK scripts, punctuation and symbols; \cref{app:fontcoverage} reports per-script coverage. Generated logits are squashed by a sigmoid and binarized at $\theta = 0.5$ to obtain output pixels.

\textbf{Stage~1 training.}
This stage uses a batch size of 384 and the AdamW optimizer \citep{loshchilov2019decoupledweightdecayregularization} to optimize \method models for 1M steps. The learning rate is linearly warmed up to 3e-4 and annealed to 3e-6 through a cosine scheduler \citep{loshchilov2017sgdrstochasticgradientdescent}. \cite{tai2024pixar} pretrained a \pixar model with 113M parameters for the generative tasks. Therefore, we used the same configuration and increased the patch size to $32 \times 32$ pixels to pretrain a \method model with 116M parameters. In addition, after testing that the maximum model that the hardware can train is around 500M, we pre-trained a \method model with 477M parameters. The stage~1 parameters can be found in~\cref{stage2hyperparameters}.

\textbf{Stage~2 training.}
The GRPO stage also uses mC4: rollouts are $K{=}3$ blocks long, sampled at temperature $1.0$ in groups of $G{=}8$ ($G{=}16$ for the 116M 32$\times$32 model); the reward of \cref{eq:reward} uses $d_{\text{blend}}\in\{0.5, 0.7\}$ and $\lambda_{\text{tile}}{=}0.7$, and the loss of \cref{eq:stage2loss} uses $\lambda_{\text{rec}}{=}0.1$ with $\beta_{\text{KL}}\in\{0.02, 0.05\}$. This stage adds only a few thousand policy steps on top of the 1M-step backbone (2{,}000 for the 8$\times$8 model, while the 32$\times$32 models iterate the recipe over up to five short generations of 1{,}500--4{,}000 steps), at constant learning rates of $10^{-5}$ or $5\times10^{-6}$. Checkpoints are selected by multilingual LAMBADA free-generation accuracy; per-model hyperparameters are reported in \cref{app:stage2rldetails}.

\subsection{Discriminative tasks}

To answer RQ1, we evaluate how \method transfers to seen and unseen languages on two multilingual classification benchmarks: SIB-200 \citep{adelani2024sib}, following \cite{kesen2025multilingual}, and XNLI \citep{conneau2018xnli}. We then address RQ3 by comparing \method models of different capacities (and patch sizes) on the English-only GLUE benchmark \citep{wang2018glue}, as in \citet{tai2024pixar}.

\subsubsection{Performance on multilingual discriminative tasks}

We divide the multilingual evaluation into \textit{seen} and \textit{unseen} languages, depending on whether the target language occurs in the pretraining data.

\begin{table*}[!t]
    \centering
    \caption{\textbf{On SIB-200, \method clearly outperforms all baselines on unseen languages, and most of all on low-resource ones, including the tokenizer-based MGPT-2 pretrained on the very same corpus.} We report the performance on both seen and unseen languages from \textbf{SIB-200}, categorized by script families. ``AVG'' denotes the average score within the respective language group, ``Non-Eng AVG'' represents the average performance excluding English, and ``All AVG'' is the mean over all 22 evaluated languages (8 seen and 14 unseen). Languages marked $^\dagger$ are low-resource (class $\leq$2 in the taxonomy of \citealp{joshi2020state}), and ``Low-res AVG'' is their mean; \cref{tab:langcodes} maps language codes to languages.}
    \setlength{\tabcolsep}{3pt}
    \scalebox{0.76}{
    \begin{tabular}{lccccccccccccccccc}
        \toprule

        \multirow{3}{*}{\textbf{Model}} & \multirow{3}{*}{\textbf{Params}} & \multirow{3}{*}{\textbf{Patch Size}} & \multicolumn{11}{c}{\textbf{Seen}} & \multicolumn{3}{c}{\textbf{Unseen}} & \\
        \cmidrule(lr){4-14} \cmidrule(lr){15-17}

        & & & \multicolumn{4}{c}{CJK} & \multicolumn{6}{c}{Latin} & \multirow{2}{*}{AVG} & \multicolumn{3}{c}{Latin} & \\
        \cmidrule(lr){4-7} \cmidrule(lr){8-13} \cmidrule(lr){15-17}

        & &                          & zh  & ja  & ko  & AVG & de  & en  & es  & it  & fr  & AVG &      & fi  & tr  & uz & \\
        \midrule

        MGPT-2           & 124M & NA & 65.9 & 81.1 & 61.1 & 69.4 & 81.7 & 87.7 & 81.7 & 81.6 & 81.2 & 82.8 & 77.8 & 44.9 & 45.2 & 40.5 &  \\
        GPT-2            & 124M & NA & 42.2 & 44.6 & 22.3 & 36.4 & 39.4 & 79.4 & 38.4 & 28.6 & 50.2 & 47.2 & 43.1 & 17.3 & 20.3 & 33.3 &  \\
        \midrule
        \pixarnospace$_{\text{stage2}}$            & 113M & 8$\times$8  & 11.0 & 25.9 & 16.6 & 17.8 & 40.6 & 81.5 & 52.3 & 53.8 & 60.5 & 57.7 & 42.8 & 29.2 & 39.7 & 42.9 &  \\
        \methodnospace$_{\text{stage1}}$ & 113M & 8$\times$8 & 58.0 & 75.3 & 38.8 & 57.4 & 78.1 & 77.9 & 79.3 & 75.3 & \textbf{83.2} & 78.8 & 70.7 & 54.4 & 53.1 & 58.2 &  \\
\methodnospace$_{\text{stage1}}$ & 116M & 32$\times$32 & 56.3 & 64.9 & 33.2 & 51.5 & 75.1 & 80.4 & \textbf{88.4} & 80.9 & 79.1 & 80.8 & 69.8 & 49.2 & 55.4 & 57.1 &  \\
\methodnospace$_{\text{stage1}}$ & 477M & 32$\times$32 & 77.1 & \textbf{79.8} & \textbf{60.2} & \textbf{72.4} & 84.9 & 79.8 & 82.8 & \textbf{81.8} & 79.5 & 81.8 & 78.3 & \textbf{58.7} & \textbf{56.7} & \textbf{59.5} &  \\
\methodnospace$_{\text{stage2}}$ & 113M & 8$\times$8 & 59.7 & 64.0 & 33.1 & 52.3 & 82.4 & 80.5 & 81.1 & 79.6 & 82.4 & 81.2 & 70.3 & 50.0 & 50.8 & 57.6 &  \\
\methodnospace$_{\text{stage2}}$ & 116M & 32$\times$32 & 56.1 & 64.1 & 28.4 & 49.5 & 74.8 & 75.8 & 81.1 & 78.2 & 78.4 & 77.7 & 67.1 & 54.5 & 56.3 & 53.7 &  \\
\methodnospace$_{\text{stage2}}$ & 477M & 32$\times$32 & \textbf{77.6} & 74.7 & 57.5 & 69.9 & \textbf{86.1} & \textbf{84.9} & 84.4 & 81.0 & 82.2 & \textbf{83.7} & \textbf{78.5} & 52.1 & 45.1 & 55.3 &  \\

        \midrule

        & & & \multicolumn{12}{c}{\textbf{Unseen (Continued)}} & \multirow{3}{*}{\shortstack{\textbf{Non-Eng}\\\textbf{AVG}}} & \multirow{3}{*}{\shortstack{\textbf{All}\\\textbf{AVG}}} & \multirow{3}{*}{\shortstack{\textbf{Low-res}\\\textbf{AVG}}} \\
        \cmidrule(lr){4-15}

        & & & \multicolumn{5}{c}{Brahmic} & \multicolumn{3}{c}{Cyrillic} & \multicolumn{3}{c}{Others} & \multirow{2}{*}{AVG} & & & \\
        \cmidrule(lr){4-8} \cmidrule(lr){9-11} \cmidrule(lr){12-14}

        & &                          & bo$^\dagger$  & bn  & hi  & ta  & te$^\dagger$  & ky$^\dagger$  & ru  & uk  & el  & he  & hy$^\dagger$  & & & & \\
        \midrule

        MGPT-2           & 124M & NA & 20.6 & 17.7 & 23.0 & 26.3 & 18.2 & 23.4 & 25.3 & 25.4 & 17.6 & 24.1 & 14.9 & 26.2 & 42.9 & 45.0 & 19.3 \\
        GPT-2            & 124M & NA & 19.9 & 20.3 & 17.0 & 20.5 & 16.9 & 24.6 & 19.5 & 20.6 & 16.5 & 16.8 & 17.0 & 20.0 & 26.0 & 28.4 & 19.6 \\
        \midrule
        \pixarnospace$_{\text{stage2}}$            & 113M & 8$\times$8 & 12.0 & 15.2 & \textbf{24.9} & 22.0 & 21.9 & 25.4 & 20.7 & 16.8 & 23.1 & 24.4 & 19.9 & 24.2 & 28.5 & 30.9 & 19.8 \\
\methodnospace$_{\text{stage1}}$ & 113M & 8$\times$8 & 21.6 & \textbf{23.1} & 19.8 & 23.1 & 23.2 & 29.8 & 31.0 & 26.6 & \textbf{32.8} & 20.5 & 15.0 & 30.9 & 43.8 & 45.4 & 22.4 \\
\methodnospace$_{\text{stage1}}$ & 116M & 32$\times$32 & 17.6 & 12.9 & 17.9 & 23.6 & \textbf{23.3} & 33.5 & 26.5 & 27.7 & 30.2 & 19.2 & 19.5 & 29.5 & 42.5 & 44.2 & 23.5 \\
\methodnospace$_{\text{stage1}}$ & 477M & 32$\times$32 & 27.1 & 20.3 & 24.0 & \textbf{25.6} & 22.1 & \textbf{37.8} & \textbf{33.8} & 27.7 & 24.1 & 23.5 & 19.8 & \textbf{32.9} & \textbf{48.0} & \textbf{49.4} & \textbf{26.7} \\
\methodnospace$_{\text{stage2}}$ & 113M & 8$\times$8 & 25.0 & 16.8 & 17.5 & 19.6 & 19.4 & 29.8 & 23.6 & 19.1 & 23.4 & \textbf{24.5} & 17.1 & 28.2 & 41.7 & 43.5 & 22.8 \\
\methodnospace$_{\text{stage2}}$ & 116M & 32$\times$32 & 25.4 & 15.3 & 20.4 & 18.5 & 18.8 & 35.3 & 26.7 & \textbf{29.8} & 29.8 & 16.6 & 17.4 & 29.9 & 41.9 & 43.4 & 24.2 \\
\methodnospace$_{\text{stage2}}$ & 477M & 32$\times$32 & \textbf{27.8} & 17.7 & 16.3 & 23.7 & 21.1 & 31.1 & 25.8 & 21.0 & 23.2 & 19.1 & \textbf{22.3} & 28.7 & 45.0 & 46.8 & 25.6 \\

        \bottomrule
    \end{tabular}
    }
    \label{tab:sib200_merged}
\end{table*}

\noindent \textbf{SIB-200 results.}
The headline result of \cref{tab:sib200_merged} answers RQ1: on the 14 languages never seen during pretraining, every \method model outperforms MGPT-2 (Unseen AVG up to 29.9 vs 26.2, and 22.4--26.7 vs 19.3 on the low-resource subset) despite being pretrained on the exact same corpus, and the largest margins appear on the scripts most distant from the pretraining mix ($+7.2$ Tibetan, $+7.7$ Kyrgyz, $+7.4$ Armenian, $+14.8$ Uzbek for the 477M stage-2 model), while \method trails only on a few languages such as Hindi and Hebrew.
On seen languages, \methodnospace$_{\text{stage2}}$ (477M) also leads on the Latin group (Seen Latin AVG 83.7 vs 82.8 for MGPT-2); the 116M model's overall seen gap (67.1 vs 77.8) stems from CJK, where a 32$\times$32 patch loses fine stroke details, and is closed entirely at 477M (Seen AVG 78.5).
A fair unseen/seen comparison with \pixelm is not possible due to its different pretraining languages \citep{kesen2025multilingual}. \cref{app:HyperparameterSIB} reports the finetuning hyperparameters.

\begin{table*}[!tb]
\centering
\caption{\textbf{477M \method outperforms the MGPT-2 baseline on seen and unseen languages on XNLI} as shown by accuracy values, especially the average non-English ones. 
} 
\scalebox{0.7}{
\begin{tabular}{lrccccccccccccccc}
\toprule
 \multirow{2}*{Models} & \multirow{2}*{Parameters} & \multirow{2}*{Patch size} & \multicolumn{6}{c}{\textbf{seen}} & \multicolumn{6}{c}{\textbf{unseen (Translate-Train)}} & \multirow{2}*{\shortstack{Non-Eng\\AVG}} \\
 \cmidrule(lr){4-9}
 \cmidrule(lr){10-15}
 & & &  de & en & es & fr & zh & AVG &   bg & el & ur & tr & ru & AVG & \\
 \cmidrule(lr){4-9}
 \cmidrule(lr){10-16}
\pixelm & 112M & 16$\times$16  & NA & NA & NA & NA & NA & NA & 52.1 & 52.9 & 47.3 & 52.2 & 38.4 & 48.6 & NA \\
MGPT-2 & 124M & NA  & 72.7 & 78.2 & 73.0 & 71.4 & 67.1 & 72.5 & 66.9 & 67.0 & 56.1 & 65.7 & 63.8 & 63.9 & 67.1  \\
BiLSTM-max & NA & NA  & 66.5 & 73.7 & 68.8 & 68.3 & 67.0 & NA & 67.4 & 66.4 & 56.6 & 64.5 & 66.5 & 64.3 & 65.8  \\
multi-BERT & 110M & NA & 75.9 & 81.9 & 77.8 & NA & 76.7 & NA & NA & NA & 61.6 & NA & NA & NA & NA  \\
\midrule

\pixarnospace$_{\text{stage2}}$ &  85M & 8$\times$8& 67.2 & 78.8 & 69.8 & 67.7 & 57.3 & NA & 60.5 & 64.6 & 50.2 & 65.5 & 64.0 & 61.0 & 63.0  \\
\methodnospace$_{\text{stage1}}$ &  113M & 8$\times$8 & 71.4 & 76.1 & 72.1 & 71.9 & 61.0 & 70.5 & 64.9 & 65.4 & 51.6 & 64.7 & 62.9 & 61.9 & 66.2  \\
\methodnospace$_{\text{stage1}}$ &  116M & 32$\times$32 & 71.8 & 75.5 & 72.7 & 72.2 & 60.3 & 70.5 & 62.8 & 64.6 & 47.0 & 65.0 & 63.2 & 60.5 & 65.5  \\
\methodnospace$_{\text{stage1}}$ &  477M & 32$\times$32 & \textbf{74.2} & \textbf{79.3} & \textbf{76.6} & 74.9 & 66.0 & \textbf{74.2} & 65.1 & \textbf{65.6} & 51.6 & 65.0 & 64.5 & 62.3 & \textbf{68.3}  \\
\methodnospace$_{\text{stage2}}$ &  113M & 8$\times$8 & 71.4 & 75.0 & 71.7 & 71.8 & 62.0 & 70.4 & 65.1 & 64.8 & 50.3 & 65.4 & 63.8 & 61.9 & 65.1  \\
\methodnospace$_{\text{stage2}}$ &  116M & 32$\times$32 & 69.8 & 75.9 & 72.9 & 72.2 & 59.0 & 70.0 & 62.9 & \textbf{65.6} & 50.3 & 65.2 & 63.9 & 61.6 & 64.6  \\
\methodnospace$_{\text{stage2}}$ &  477M & 32$\times$32 & 73.8 & 79.0 & 75.9 & \textbf{75.2} & \textbf{66.6} & 74.1 & \textbf{65.7} & \textbf{65.6} & \textbf{51.8} & \textbf{66.4} & \textbf{64.6} & \textbf{62.8} & 67.3  \\

\bottomrule
\end{tabular}
}

\label{XNLI1M}
\end{table*}

\noindent \textbf{XNLI results.}
We select ten XNLI languages, five \textit{seen} (German, English, Spanish, French, Chinese) and five \textit{unseen} (Bulgarian, Greek, Urdu, Turkish, Russian), and follow the \textit{translate-train} paradigm: each model is finetuned on machine-translated training data of the target language, enabling a fair comparison with \texttt{multi-BERT} and the \texttt{BiLSTM-max} baseline of \citet{conneau2018xnli}.

As shown in \cref{XNLI1M}, \method is on par with or better than MGPT-2 throughout: \methodnospace$_{\text{stage2}}$ (477M) leads on seen languages (Seen AVG 74.1 vs 72.5) and on the overall Non-Eng AVG (67.3, the best in the table), and on unseen languages it stays within about a point of the best text-based baselines (62.8 vs 63.9--64.3). \method also clearly outperforms the pixel-based alternatives: \pixelm on every unseen language, and the English-only \pixar by over 4 points on the Non-Eng AVG. \cref{HyperparameterXNLI} lists the finetuning hyperparameters.

The contrast between the two benchmarks is informative: besides containing no low-resource language (all are class $\geq$3 in the taxonomy of \citealp{joshi2020state}), the unseen XNLI set (Bulgarian, Greek, Turkish, Russian, Urdu) is dominated by European languages sharing alphabets and cognates with our pretraining mix, so much of a subword vocabulary carries over and tokenizer-based models like MGPT-2 generalize to them at little cost; moreover, its Translate-Train protocol supplies ${\sim}$393k translated examples per language, so much supervision that a from-scratch BiLSTM tops the column (64.3); the benchmark thus rewards supervised capacity more than pretraining transfer. SIB-200, with only ${\sim}$700 training sentences per language and unseen languages that are genuinely distant in family, script and culture, removes all three crutches, and this is exactly where the pixel interface pulls ahead. It is thus the pixel interface, not the data recipe, that carries transfer to truly cross-script, low-resource settings.

\subsubsection{Performance on English-only discriminative tasks}

\begin{table*}[!t]
\centering
\caption{\textbf{\method performs better than \pixar and MGPT-2 on the GLUE benchmark} as shown by Matthew's correlation for COLA, Spearman's $\rho$  for STSB and F1 score values for MRPC and QQP, while other tasks used accuracy. The number below the each tasks denotes the amount of training samples per task.} 
\setlength{\tabcolsep}{2pt}
\scalebox{0.9}{
\begin{tabular}{lrccccccccccc}
\toprule
\multirow{2}*{Models} & \multirow{2}*{Parameters} & Patch size& MNLI-m/mm &  QQP & QNLI & SST-2 & COLA & STSB & MRPC & RTE & WNLI & \multirow{2}*{AVG}\\
\cmidrule(lr){4-12}
 &  & (pixel) & 392k  &  363k & 108k & 67k  & 8.5k & 5.7k & 3.5k & 2.5k & 635 &  \\
\midrule 
GPT-2 & 126M & NA & 81.0 & 89.4 & 87.7 & 92.5 & 77.0 & 74.9 & 71.5 & 52.0 & 54.9 & 75.6  \\
MGPT-2 & 126M & NA & 78.2/78.5 & 85.2 & 85.3 & 87.9 & 33.7 & 87.0 & 82.3 & 64.6 & 57.4 & 74.0  \\
BERT & 110M & NA &  84.0/84.2 & 87.6 & 91.0 &  92.6 &  60.3 & 88.8 & 90.2 & 69.5 & 51.8 & 80.0  \\
\midrule

\pixel                & 86M  & 16$\times$16                       & 78.1/78.9 & 84.5          & \textbf{87.8} & 89.6  & 38.4          & 81.1          & \textbf{88.2} & 60.5 & 53.8          & 74.1                 \\

\pixarnospace$_{\text{stage1}}$ & 85M & 8$\times$8 &  78.4/78.6 & 85.6  & 85.7 & 89.0  & \textbf{39.9} & 81.7 & 83.3 & 58.5 & 59.2 & 74.0  \\
\pixarnospace$_{\text{stage2}}$ & 85M & 8$\times$8 & 79.7/80.1 &  86.3 & 85.7 &  89.3 &  37.0 & 82.4 & 82.8 &  57.7 & \textbf{60.6} & 74.2   \\
\midrule

\methodnospace$_{\text{stage1}}$ & 113M & 8$\times$8 & 75.7/75.3 & 85.3 & 84.3 & 88.4 & 24.2 & 85.1 & 85.8 & 62.5 & 56.3 & 72.3  \\
\methodnospace$_{\text{stage1}}$ & 116M & 32$\times$32 & 75.7/75.7 & 85.0 & 83.9 & 87.7 & 17.6 & 84.6 & 85.6 & 62.1 & 56.3 & 71.4  \\
\methodnospace$_{\text{stage1}}$ & 477M & 32$\times$32 & 79.7/80.1 & \textbf{87.4} & 87.4 & 89.4 & 29.5 & \textbf{86.2} & 86.2 & \textbf{63.9} & 56.3 & \textbf{74.6}  \\

\methodnospace$_{\text{stage2}}$ & 113M & 8$\times$8 & 75.3/76.0 & 85.4 & 83.8 & 89.1 & 20.1 & 85.4 & 81.5 & 63.2 & 56.3 & 71.6  \\
\methodnospace$_{\text{stage2}}$ & 116M & 32$\times$32 & 75.4/75.8 & 85.3 & 84.4 & 88.2 & 20.3 & 84.6 & 85.6 & 62.1 & 59.2 & 72.1  \\
\methodnospace$_{\text{stage2}}$ & 477M & 32$\times$32 & \textbf{79.8/80.0} & 87.2 & 87.2 & \textbf{89.9} & 29.6 & 86.1 & 86.5 & 52.7 & 59.2 & 73.8  \\
\bottomrule

\end{tabular}
}

\label{GLUE1Mtable}  
\end{table*}

We finetune \method with a newly initialized prediction head per task; for sentence-pair tasks, a black patch separates the two sentences and the embedding of the last black patch feeds the head, a strategy that proved effective in previous work \citep{rust2022language, tai2024pixar}. We reuse \pixar's finetuning hyperparameters and early-stopping criteria, tuning only the learning rate per task, as small datasets are sensitive to it \citep{tai2024pixar} (\cref{HyperparameterGLUE}).

To address RQ3, we first evaluate \method in English-only settings using the GLUE benchmark. As shown in Table \ref{GLUE1Mtable}, performance improves consistently with capacity (71.4 to 74.6 average from 116M to 477M), and the 477M \method with 32$\times$32 patches matches or exceeds the 8$\times$8 \pixar on most tasks (e.g., $+8.3$ on RTE), showing that the larger patch size does not compromise language understanding. \cref{GLUE100ktable} reports the same comparison at 0.1M pretraining steps.

The largest \method (477M) also significantly narrows the gap with tokenizer-based models such as GPT-2. It still lags behind BERT, but \method is a fully autoregressive, multilingual model, whereas BERT is an encoder-only architecture limited to classification.

\subsection{Generative tasks}

We selected bAbI \citep{weston2015towards} and LAMBADA \citep{paperno2016lambada} as generative tasks for our evaluation (RQ2). We additionally probe CJK free generation on a machine-translated bAbI. For the LAMBADA tasks, Huggingface provides machine-translated versions for \href{https://huggingface.co/datasets/EleutherAI/lambada_openai}{German, Spanish, French, and Italian}. We used accuracy to measure how well the model predicted the outcome of the generation task. 
During evaluation, we employed Qwen3.6-35B-A3B as an OCR model to recognize the text within the generated patches. To prevent the model from independently inferring the correct answers, we restricted its input to the generated patches and only the tail of the prompt (\cref{app:ocrjudge}).

Table \ref{babilamb} demonstrates the performance of \pixar and \method on bAbI and LAMBADA for languages with an alphabetic writing system. Comparing the two stage-1 models of comparable size (113M with 8$\times$8 patches and 116M with 32$\times$32 patches), we observe a decrease in LAMBADA performance as patch size increases, since a larger patch forces the model to commit to more characters at once. Stage~2 is what turns these backbones into usable generators: it multiplies LAMBADA accuracy by $1.7$--$2.3\times$ at every scale (6.0$\rightarrow$10.4, 4.7$\rightarrow$10.7 and 7.2$\rightarrow$15.3) and nearly doubles bAbI accuracy for the 477M model (18.8$\rightarrow$34.6).
Furthermore, since English was the only language used for training, \pixar outperforms \method with a similar number of parameters (116M) for English tasks.
The CJK half of \cref{babilamb} tells a different story. Stage~2 improves every language except Chinese: Japanese rises from 2.0 to 6.4 at 116M and from 6.8 to 11.0 at 477M, Korean from 2.2 to 5.8 and from 0.4 to 1.0, while Chinese slips slightly at both scales. All CJK scores nevertheless remain far below the English ones. Two properties of these scripts make the task harder. Their glyphs are visually intricate, so a patch has to be reproduced with far less pixel redundancy than a Latin letter. They are also much denser: \cref{tab:tokenperpatch} reports how much text one rendered patch carries in each language. Across the five Latin-script languages a patch holds about $1.6$ characters, worth between $0.42$ and $0.62$ GPT-2 tokens. A Chinese patch holds a single character worth $1.8$ tokens and a Korean patch $1.2$ characters worth $2.3$ tokens. Emitting one CJK patch is therefore committing to three to five times as much text as emitting one English patch, in a single step and without a tokenizer to fall back on. The tokenizer baselines do not solve the task either: on the same prompts they stay at or below the accuracy of always answering the most frequent location, which is $18.8$.

\begin{table}[h]
\centering
\caption{\textbf{A CJK patch carries three to five times more text than an English one.} Characters and GPT-2 tokens contained in a single rendered $32{\times}32$ patch, measured on mC4 samples of the eight pretraining languages. MGPT-2 shares the GPT-2 tokenizer, so its token counts are identical.}
\begin{tabular}{lcccccccc}
\toprule
 & en & de & es & fr & it & zh & ja & ko \\
\midrule
characters per patch & 1.59 & 1.58 & 1.59 & 1.60 & 1.66 & 1.02 & 1.01 & 1.24 \\
GPT-2 tokens per patch & 0.42 & 0.62 & 0.58 & 0.56 & 0.58 & 1.77 & 1.27 & 2.29 \\
\bottomrule
\end{tabular}

\label{tab:tokenperpatch}
\end{table}

For the other languages, \pixar also has a competitive performance. Presumably, this is because they also use a Latin script. After increasing the number of model parameters to 477M, \method outperforms \pixar on the English bAbI task and has a similar performance on the LAMBADA task. For all languages, increasing the number of parameters improves accuracy. We deduce that additional parameters are required to capture the specifics of each language, providing preliminary evidence that further scaling the model, both in terms of parameter size as well as training it with more scalable training regimes, could lead to greater performance.

\begin{table*}[!t]
\centering
\caption{\textbf{Scaling \method to 477M outperforms the English-only \pixar and text-based GPT-2 on average across multilingual generative tasks.} 
For bAbI, we report few-shot free-generation accuracy (dash-delimited answers; 2 shots for the 477M model, 3 for the others) and for LAMBADA zero-shot last-word free-generation accuracy. \method{} generations are judged by a Qwen3.6-35B OCR judge with a pixel-exact shortcut and confusable folding, reading only the generated region on bAbI and the generation with a short anchoring context on LAMBADA, so that the judge never sees the question it is transcribing an answer to (\cref{app:ocrjudge}); the $stage2\text{-}gan$ row is the adversarial stage-2 counterpart of the 8$\times$8 model (\cref{app:stage2ablation,app:stage2rldetails} for training details). \pixar{} numbers are from \citet{tai2024pixar}.} 
\scalebox{0.86}{
\begin{tabular}{ccccccccccccc}
\toprule
Models & \#Params & Patch & \multicolumn{4}{c}{bAbI} &
\multicolumn{6}{c}{LAMBADA} \\
\cmidrule(lr){4-7}
\cmidrule(lr){8-13}
&&size&en & zh & ja & ko & en & de & es &fr & it & avg \\
\midrule

GPT-2 & 124M  & -  &  26.8 & 17.4 &  17.4 &  25.6 & 17.1  & 4.3 &  6.1 & 7.9 &  6.3 & 8.3 \\
MGPT-2 & 124M & -  &  24.3 & 17.6 &  21.8 &  10.6 & 18.8  & 18.3 &  16.0 & 20.6 &  17.3 & 18.2  \\
\midrule

 \pixarnospace$_{\text{stage2}}$ & 113M & 8$\times$8 &  19.6 &  0.0 &   0.0 &   0.0 &  \textbf{13.8} &  3.6 &  2.5 &  5.9 &  5.6 & 6.3  \\

\methodnospace$_{\text{stage1}}$ & 113M & 8$\times$8 &  11.4 &  0.0 &   0.0 &   0.0 &  3.8 &  10.8 &  4.6 &  5.0 &  5.8 & 6.0 \\
\methodnospace$_{\text{stage1}}$ & 116M & 32$\times$32 &  15.2 &  3.8 &   2.0 &   2.2 &  2.4 &  8.6 &  3.8 &  4.2 &  4.6 & 4.7 \\
\methodnospace$_{\text{stage1}}$ & 477M & 32$\times$32 &  18.8 &  1.8 &   6.8 &   0.4 &  5.2 &  12.0 &  6.4 &  5.0 &  7.4 & 7.2 \\

\methodnospace$_{\text{stage2}}$ & 113M & 8$\times$8 &  15.6 &  0.0 &   0.0 &   0.0 &  8.8 &  9.8 &  12.4 &  9.6 &  11.4 & 10.4 \\
\methodnospace$_{\text{stage2}}$ & 116M & 32$\times$32 &  16.6 &  3.2 &   6.4 &   5.8 &  9.2 &  12.6 &  11.6 &  10.2 &  9.8 & 10.7 \\
\methodnospace$_{\text{stage2}}$ & 477M & 32$\times$32 &  \textbf{34.6} &  1.4 &  \textbf{11.0} &   1.0 &  13.4 &  \textbf{16.4} &  \textbf{17.6} &  \textbf{14.0} &  \textbf{15.0} & \textbf{15.3} \\
\methodnospace$_{\text{stage2-gan}}$ & 113M & 8$\times$8 &  7.8 &  0.0 &   0.0 &   0.0 &  4.6 &  11.2 &  4.6 &  4.8 &  5.4 & 6.1 \\

\bottomrule
\end{tabular}
}

\label{babilamb}
\end{table*}

\begin{figure*}[!t]
    \centering
    \includegraphics[width=\textwidth]{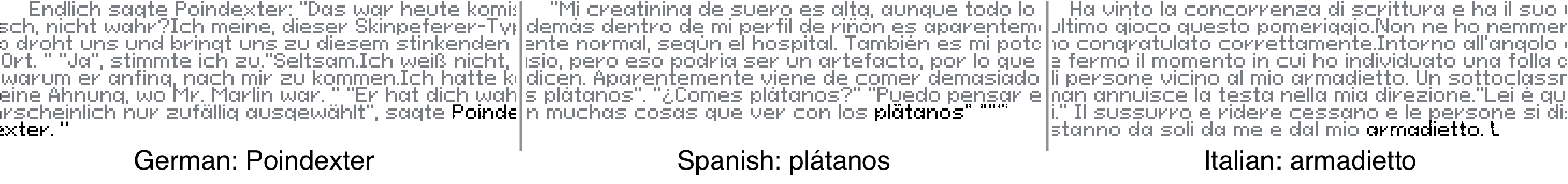}
    \caption{\textbf{\method can handle multilingual text as image}, as shown here for examples of correct completions (black) for German, Spanish and Italian prompts (gray) on LAMBADA.}
    \label{lamb3}
\end{figure*}

Figure \ref{lamb3} shows the correct examples for German, Spanish and Italian Lambada samples (see Figure \ref{lamb116good} for more correct examples from LAMBADA). Besides, figure \ref{lamb116bad} provided wrong Lambada examples and figure \ref{babi116good} and Figure \ref{babi116bad} shows correct and wrong answers of bAbI.

\subsection{Robustness to input perturbations}

It is well known that deep learning models for vision are sensitive to adversarial attacks~\citep{hendrycks2019benchmarking}, which is a concern that carries over to pixel-based language models. To answer RQ4 concerning robustness against visual attacks, we focus on orthographic attacks, which replace individual letters with visually similar ones.

Following \citet{tai2024pixar}, who use the method of \citet{eger2020hero}, we attack English LAMBADA and bAbI prompts by replacing each letter, with probability equal to the attack ratio, by a visually similar character from a \href{https://util.unicode.org/UnicodeJsps/confusables.jsp}{Unicode Technical Standard \#39} confusable set. We restrict the substitutes to characters that our bundled fonts render as true lookalike glyphs (\cref{app:fontcoverage}), and evaluate the stage-2 \method models under the free-generation protocol of \cref{babilamb}; the tokenizer baselines (MGPT-2, pretrained on the same corpus as \method, and the original English GPT-2) receive exactly the same attacked text.

\begin{figure*}[!t]
\begin{minipage}{.65\textwidth}
    \centering
    \includegraphics[width=1\textwidth]{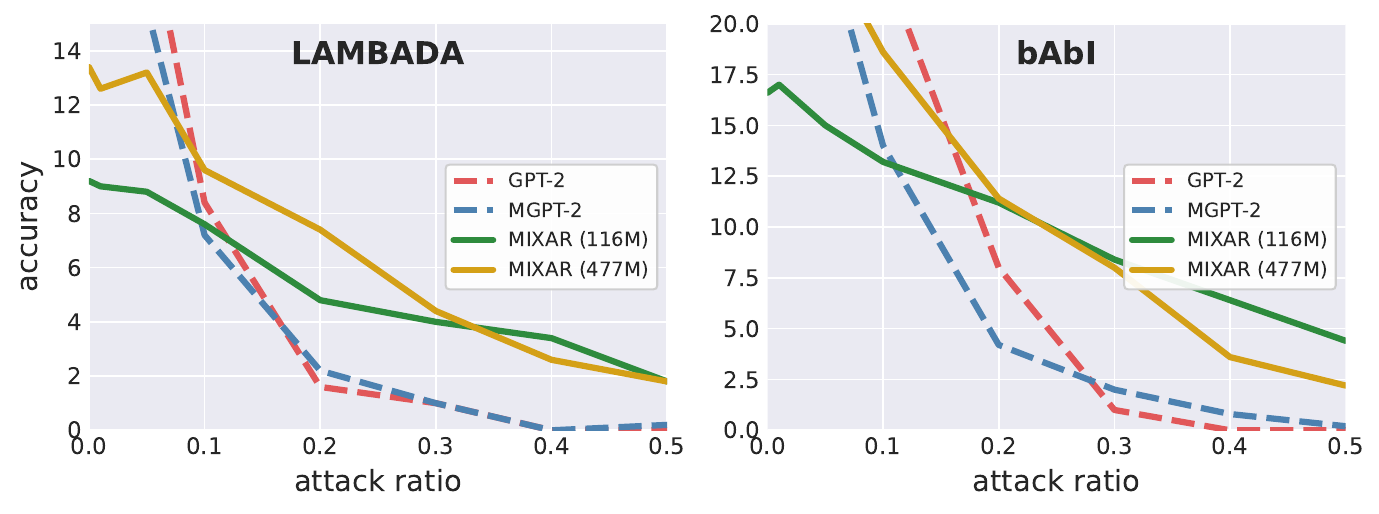}
\end{minipage}\hfill\begin{minipage}{.3\textwidth}
        \caption{\textbf{\method withstands English orthographic attacks more gracefully than tokenizer-based models}: accuracy under increasing attack ratios on LAMBADA and bAbI (free-generation protocol, lookalike glyphs from the bundled fonts). Attack examples are shown in \cref{confuse_mixar}.
        }
    \label{babilambconfuse}
\end{minipage}
\end{figure*}

\begin{figure}[!t]
    \centering
    \includegraphics[width=\textwidth]{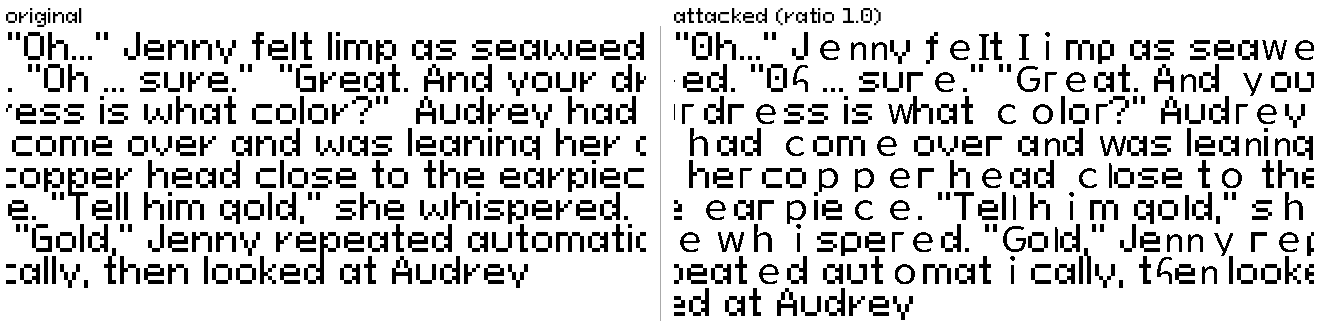}
    \caption{\textbf{An orthographic attack barely changes the rendered image.} The same English LAMBADA prompt rendered before (left) and after (right) an attack at ratio $1.0$, which replaces every attackable character by a Unicode lookalike. The text a tokenizer reads is almost entirely different, while the pixels \method reads are nearly unchanged.}
    \label{attackexample}
\end{figure}

Figure~\ref{babilambconfuse} (per-ratio numbers in \cref{confuse}) shows the same picture on both tasks and against both tokenizer baselines: MGPT-2, which shares our multilingual pretraining corpus, and the original English GPT-2 start from higher clean accuracies but collapse quickly, while \method degrades gracefully. The curves cross between ratios of 0.05 and 0.2, and beyond that the pixel models dominate: at ratio 0.2 the 477M \method retains 11.4 on bAbI against 8.0 and 4.2 for the two baselines, and from ratio 0.4 both baselines are at or near zero on both tasks while \method still answers correctly several percent of the time. At a ratio of 0.1, \method models retain 54--83\% of their clean accuracy, against 26--56\% for the tokenizer-based models. \cref{attackexample} shows why: at the extreme of the attack the rendered image is still almost the original one, while the underlying codepoint sequence has been almost entirely replaced.

\section{Conclusion}

We introduced \method, the first multilingual autoregressive pixel-based language model, pretrained on eight high-resource languages spanning Latin and CJK scripts and refined with a GRPO-based free-generation stage.
Our experiments answer the four questions we set out with.
Pretrained on the same corpus as its tokenizer-based counterpart, \method transfers better to unseen low-resource languages, with its largest margins on the scripts most distant from the pretraining mix (RQ1), a step toward more equitable language technology. 
The GRPO stage proves both more stable than adversarial training and substantially more effective, multiplying free-generation accuracy by up to $2.3\times$ on multilingual LAMBADA and nearly doubling bAbI accuracy at 477M, where its adversarial counterpart stalls at the backbone's level (RQ2).
Pixel-based models scale: every benchmark improves monotonically from 116M to 477M parameters, and the larger model closes the CJK gap to tokenizer-based baselines on seen languages (RQ3).
Finally, under orthographic attacks composed of lookalike glyphs, \method degrades far more gracefully than its tokenizer-based counterpart as the attack intensifies (RQ4), supporting the view that pixel-based models offer a viable path toward more script-agnostic and tokenizer-free language processing~\citep{kharitonov2022textless}.

At the same time, our results highlight important open challenges.
Although \method generalizes encouragingly to unseen languages, performance remains uneven, particularly for scripts that differ substantially from those observed during training.
Its visual robustness is also bounded by the font it was trained on: attacks that substitute lookalike glyphs drawn from fonts unseen during pretraining, or rendering the input in an unseen font altogether, degrade \method sharply (\cref{sec:font_variations}).
As in \method each pixel is conditionally independent from the others, we plan to retrieve dependencies to improve generation \citep{grivas2025fast}.
To this end, we can replace our final layer with a circuit \citep{choi2020probabilistic} or a diffusion model \citep{croitoru2023diffusion}, aiming for broader language coverage and improved robustness to visual variations in the input, such as font changes.
Overall, our findings suggest that multilingual pixel-based language models are a promising direction for building more inclusive and resilient language technologies.

\clearpage
\section*{Reproducibility Statement}

We commit to releasing every artifact required to reproduce the results in this paper. Below we summarize what will be released and where each ingredient is specified.

\noindent \textbf{Code and artifacts.} We release:
\begin{itemize}
    \item our training framework, together with all training and evaluation scripts and the launch YAML configuration files that reproduce every experiment reported in this paper, from pretraining to finetuning, stage-2 GRPO and all evaluations (\codeurl);
    \item the deterministic text-rendering tool as a standalone package, bundling the exact fonts used in all experiments (\renderurl);
    \item all pretrained checkpoints on the Hugging Face hub, including the three stage-1 backbones, the three stage-2 models, and the MGPT-2 baseline pretrained on the same corpus (\modelurl).
\end{itemize}

\noindent \textbf{Models and training.} 
All architecture, rendering and stage-1 pretraining hyperparameters are listed in \cref{stage2hyperparameters}, and the training objectives are defined in \cref{sec:multilingual}.
For the stage-2 GRPO refinement, the reward, loss and gating mechanism are specified in \cref{eq:reward,eq:stage2loss} and the surrounding text, and every per-model hyperparameter, including group sizes, reward blends, gating thresholds, learning rates and the checkpoint-selection protocol, is tabulated in \cref{app:stage2rldetails}.
Finetuning hyperparameters are reported in \cref{HyperparameterGLUE,HyperparameterXNLI,HyperparameterSIB}.

\noindent \textbf{Data and rendering.}
Both training stages sample from the publicly available mC4 corpus \citep{raffel2020exploring}; the eight-language composition is reported in \cref{tab:dataset_stats}, and the sampling and filtering procedure is fully described in \cref{sec:mutli-pre-train}. The fonts and their per-script coverage are documented in \cref{app:fontcoverage,alphabet}.

\noindent \textbf{Evaluation.}
All benchmarks are publicly available on Hugging Face, and our codebase links the exact versions. The free-generation scoring protocol, including the pixel-exact shortcut and the OCR judge, is fully specified in \cref{app:ocrjudge}, and the orthographic-attack methodology follows \citet{tai2024pixar} and is described in \cref{sec:exps}, enabling independent robustness testing.

\noindent \textbf{Computational resources.}
\begin{itemize}
    \item Stage-1 pretraining (1M steps each) took ${\sim}$29 hours for the 113M 8$\times$8 model, ${\sim}$31 hours for the 116M 32$\times$32 model, and ${\sim}$62 hours for the 477M model, on four NVIDIA H200 GPUs.
    \item The stage-2 GRPO refinement is comparatively cheap: one generation (1{,}500--4{,}000 policy steps) takes 1--2.5 hours on four H100 80GB GPUs, and the released models used between one and five generations. Checkpoint selection adds one LAMBADA evaluation per candidate, 15--25 minutes on a single GPU with a co-located OCR judge.
    \item Finetuning is lighter still: 4.4 H200 GPU-hours for the 22 SIB tasks (116M) and 48.7 H200 GPU-hours for the 10 XNLI tasks (477M), with the 5 LAMBADA evaluations taking about 25 minutes on 2 H200s.
\end{itemize}

\clearpage

\bibliography{main}
\bibliographystyle{tmlr}

\appendix
\crefalias{section}{appendix}
\crefalias{subsection}{subappendix}
\clearpage

\label{sec:appendix}
\section{Letters not available in the English alphabet}
\begin{table*}[htbp]
\centering
\caption{This table shows the letters outside the English alphabet for other 4 latin languages of pretraining dataset. }
\scalebox{0.8}{
\begin{tabular}{c|c}
\hline
language & Letters outside the English alphabet\\
\hline
German & (Ä	ä) (Ö ö) (ẞ ß) (Ü ü) \\
French & (À à) (Â â) (Æ æ) (Ç ç) (É	é) (È è) (Ê ê) (Ë ë) (Î î)	(Ï ï)(Ô	ô) (Œ œ) (Û	û) (Ù ù) (Ü ü) (Ÿ ÿ) \\
Spanish &  (Á	á)	(Ch	ch)	(É	é)	(Í	í)	(Ll	ll)	(Ó	ó) (Ú	ú)	(Ü	ü) \\
Italian & (À à)	(È è)	(É é)	(Ì ì)	(Í í)	(Î î)	(Ò ò)	(Ó ó)(Ù ù)	(Ú ú)	 \\
\hline
\end{tabular}
}

\label{alphabet}  
\end{table*}

Table \ref{alphabet} shows the letters outside the English alphabet of German, French, Spanish and Italian which are the languages chosen for \method pretraining. This alphabet demonstrates the similarity and difference between the five languages in terms of letters. Because \method is a visual language model, similar letters make their text patches similar. 

\section{Increasing patch size}
\label{sec:patch-size}

\begin{figure*}[htp]
    \centering
    \includegraphics[width=15cm]{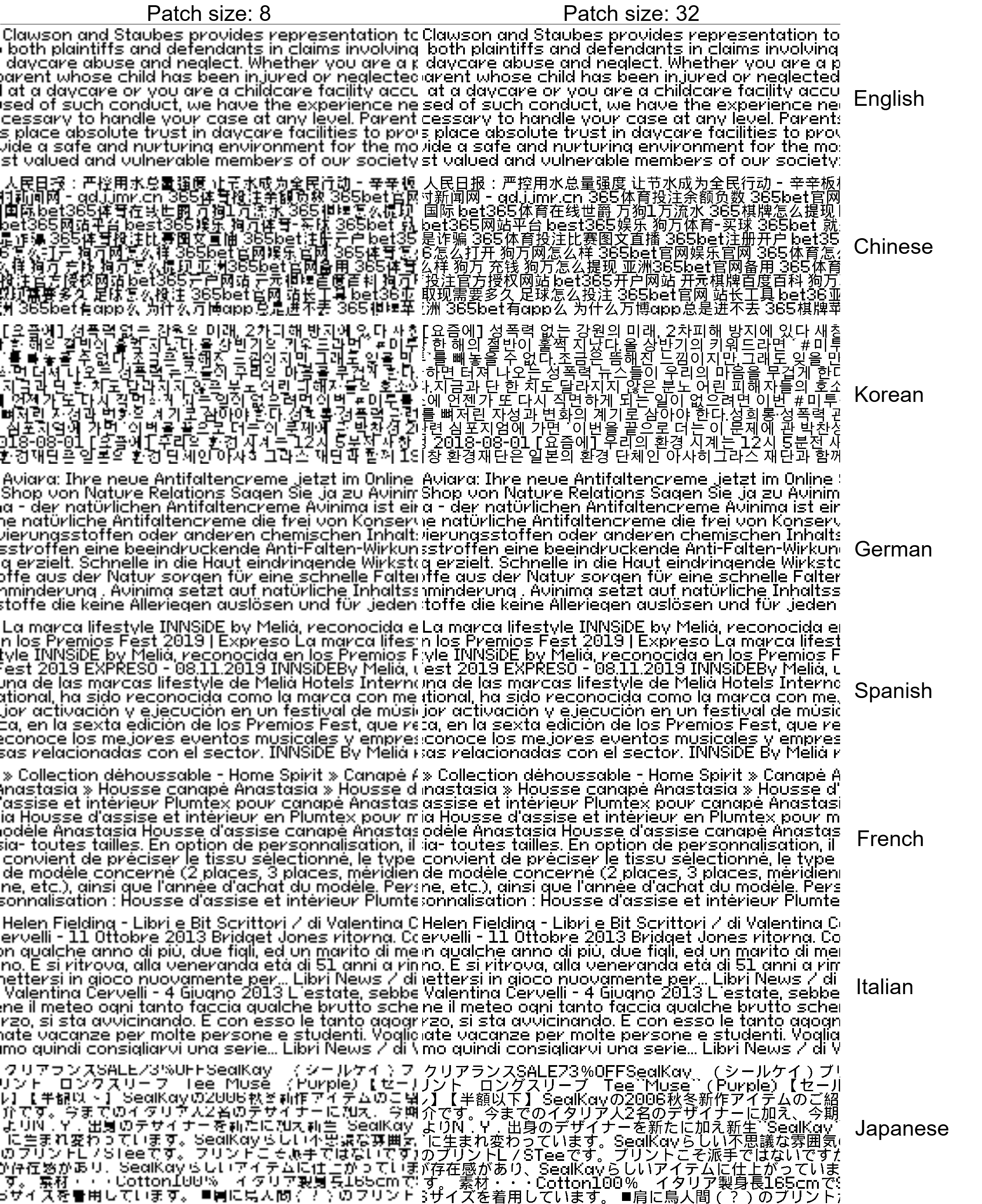}
    \caption{This image shows the comparison of patch size 8 and 32 for all pretraining scripts.}
    \label{8samples}
\end{figure*}

Previous pixel-based language models, such as \pixar, chose a patch size of up to 8$\times$8 pixels to represent the rendered text. 
This choice was made to trade off computation with generation quality, as generating several
coherent new patches of pixels is more challenging than just predicting a single token.
This was sufficient because their model was trained and evaluated only on  English, where characters can be easily rendered and read on 8$\times$8 pixel patches. 
However, this choice becomes a limiting factor when one wants to extend pixel-based models to non-Latin scripts. 
In fact, the pretraining dataset of our \method contains languages such as Chinese, Korean, and Japanese, whose characters require a higher resolution to be correctly disambiguated, as shown in~\cref{8samples}. 
For this reason, we go beyond what was studied \cite{lotz2023text} and increase the patch size to 32$\times$32 pixels as an essential step to correctly represent these languages. 
While this facilitates encoding more complex scripts, it also increases the complexity of training due to the increased image resolution. 

\section{Language codes}
\label{app:langcodes}

\cref{tab:langcodes} lists every language code used in our evaluation tables,
together with its script and whether the language occurs in the \method
pretraining corpus. Languages marked $^\dagger$ are low-resource (class
$\leq$2 in the taxonomy of \citealp{joshi2020state}).

\begin{table}[htbp]
\centering
\caption{Language codes used in the evaluation tables, with script and
pretraining status. $^\dagger$: low-resource languages (class $\leq$2,
\citealp{joshi2020state}).}
\scalebox{0.85}{
\begin{tabular}{llll}
\toprule
Code & Language & Script & Pretraining \\
\midrule
de & German & Latin & seen \\
en & English & Latin & seen \\
es & Spanish & Latin & seen \\
fr & French & Latin & seen \\
it & Italian & Latin & seen \\
ja & Japanese & Kana + Han & seen \\
ko & Korean & Hangul & seen \\
zh & Chinese & Han & seen \\
\midrule
bg & Bulgarian & Cyrillic & unseen \\
bn & Bengali & Bengali & unseen \\
bo$^\dagger$ & Tibetan & Tibetan & unseen \\
el & Greek & Greek & unseen \\
fi & Finnish & Latin & unseen \\
he & Hebrew & Hebrew & unseen \\
hi & Hindi & Devanagari & unseen \\
hy$^\dagger$ & Armenian & Armenian & unseen \\
ky$^\dagger$ & Kyrgyz & Cyrillic & unseen \\
ru & Russian & Cyrillic & unseen \\
ta & Tamil & Tamil & unseen \\
te$^\dagger$ & Telugu & Telugu & unseen \\
tr & Turkish & Latin & unseen \\
uk & Ukrainian & Cyrillic & unseen \\
ur & Urdu & Perso-Arabic & unseen \\
uz & Uzbek & Latin & unseen \\
\bottomrule
\end{tabular}
}

\label{tab:langcodes}
\end{table}

\section{Font character coverage}
\label{app:fontcoverage}

\cref{tab:fontcoverage} reports, per Unicode block, how many codepoints are
covered by the primary pixel font and by the bundled Go~Noto~Current fallback
(two further bundled fonts, Noto Sans Symbols~2 and Noto Emoji, extend symbol
and emoji coverage and are omitted for brevity). The renderer resolves every
glyph against this fixed, bundled set only --- it never falls back to fonts
installed on the host --- so renderings are reproducible across machines, and
any codepoint outside the bundled set is reported explicitly instead of being
silently substituted. The uncovered part of the CJK Unified Ideographs block
consists predominantly of rare and historic ideographs; all Hangul syllables
and kana are covered in full.

\begin{table}[htbp]
\centering
\caption{Per-block character coverage (covered/total codepoints) of the two
main bundled fonts of our rendering tool. The primary PixeloidSans font covers
Latin scripts; Go~Noto~Current supplies the remaining scripts, including full
kana and Hangul coverage and the common CJK ideographs.}
\scalebox{0.85}{
\begin{tabular}{lrr}
\toprule
Unicode block & PixeloidSans & Go Noto Current \\
\midrule
Basic Latin (ASCII) & 95/128 & 98/128 \\
Latin-1 Supplement & 117/128 & 96/128 \\
Latin Extended-A & 128/128 & 128/128 \\
Latin Extended-B & 97/208 & 208/208 \\
Greek and Coptic & 111/144 & 135/144 \\
Cyrillic (+ Supplement) & 174/304 & 304/304 \\
Armenian & 1/96 & 91/96 \\
Hebrew & 0/112 & 87/112 \\
Arabic & 0/256 & 255/256 \\
Devanagari & 0/128 & 128/128 \\
Bengali & 1/128 & 96/128 \\
Thai & 1/128 & 87/128 \\
General Punctuation & 38/112 & 111/112 \\
CJK Symbols and Punctuation & 0/64 & 64/64 \\
Hiragana & 0/96 & 93/96 \\
Katakana & 0/96 & 96/96 \\
Hangul Syllables & 0/11172 & 11172/11172 \\
CJK Unified Ideographs & 2/20992 & 11928/20992 \\
\bottomrule
\end{tabular}
}

\label{tab:fontcoverage}
\end{table}

\section{Font variations} \label{sec:font_variations}

In addition to orthographic attacks, we also evaluate whether \method is robust to changes in the font used. Specifically, we select the babi task as reference, and test the performance of the fine-tuned model with the unseen font \href{https://fontmeme.com/ziti/smalltalk-sans10-font/}{Smalltalk Sans10} and \href{https://fontmeme.com/fonts/tiny-unicode-font/}{Tiny Unicode}. The reason for this outcome might be that the dataset used by Mixar for pre-training only employed one type of font during the rendering process. This makes it very difficult for the model to understand the patches after the font has been changed. According to our Orthographic attacks experiment, the performance of \method drops to almost zero when the attack ratio is 0.5. Changing the font is more like conducting an orthographic attack on all the letters of the prompt.

\section{Detailed pretraining hyperparameters}

\begin{table*}[htbp]
\centering
\caption{Rendering configuration, model structure and stage-1 pretraining hyperparameters of the 116M / 477M \method models. Stage-2 (GRPO) hyperparameters are reported separately in \cref{stage2rlhp}.}
\scalebox{0.6}{
\begin{tabular}{cc|cc|cc}
\hline
 \multicolumn{2}{c}{Render Configuration}& \multicolumn{2}{|c|}{Model Structure} & \multicolumn{2}{c}{Stage 1 Hyperparameters}\\
\hline
patch length & 2 & layers & 12 / 24 & peak lr & 3e-4 \\
patch number & 360 & attention heads & 12 / 14 & min. lr &  3e-5 \\
render DPI & 80 & hidden size & 768 / 896& lr scheduler& CosineAnnealing \\
font size & 32 & activation &  SwiGLU&optimizer & AdamW \\
patch size & 32 & intermediate size &3072 / 6656 & $\beta_{1}$ & 0.9 \\
font & PixeloidSans & parameters & 116M / 477M& $\beta_{2}$& 0.95 \\
binary & true &  & & weight decay&0.1 \\
Temperature (T) & 1 &  & & steps& 1M \\
Threshold ($\theta$) & 0.5 &  & &warm up & 2000 \\
& &  & &batch size & 384 \\
& &  & &precision & fp16 \& fp32 \\
&    & & & random seed & 42 \\
\hline
\end{tabular}
}

\label{stage2hyperparameters}  
\end{table*}

To ensure reproducibility of our pretraining runs, we report in 
\cref{stage2hyperparameters} the hyperparameters used on rendering, model structuring and stage-1 training for \method; the stage-2 GRPO hyperparameters are in \cref{stage2rlhp}.

\section{Hyperparameters for GLUE}

\cref{HyperparameterGLUE} provides the hyperparameters applied in the GLUE benchmark.

\section{Hyperparameters for XNLI}

\cref{HyperparameterXNLI} shows the hyperparameters used in the XNLI benchmark.

\section{Hyperparameters for SIB}
\label{app:HyperparameterSIB}
\cref{HyperparameterSIB} shows the hyperparameters used in the SIB benchmark.

\section{Stage-2 objective ablation on the 8$\times$8 model}
\label{app:stage2ablation}

The 8$\times$8 \method rows in the main tables use a GRPO-style stage-2: the same
per-patch discriminator (conditioned on the real prefix) is used as a reward
model, with a group-relative, std-normalized advantage on sampled patches,
a KL anchor to the frozen stage-1 model, and a reconstruction anchor. Here we
compare it against adversarial (GAN) stage-2 training under the same
discriminator, data and budget regime. \cref{stage2ablate} reports
teacher-forced LAMBADA last-word exact-match (all pixels of the last word
correct) on 2{,}500 passages (500 per language).

The raw non-saturating objective ($\mathcal{L}_{\text{adv}} + \lambda_m \mathcal{L}_{\text{rec}}$)
diverges for all $\lambda_m \in \{1, 3, 9\}$: the adversarial pull is a
constant-magnitude force while the reconstruction gradient vanishes on the data
manifold, so the generator drifts off-manifold at a rate $\propto 1/\lambda_m$.
Adding the adaptive gradient-ratio weight of \citet{tai2024pixar} (normalizing
the adversarial gradient to the reconstruction gradient's norm at the
generator's last layer) stabilizes every configuration we tried — across
$\lambda$, D/G schedules (threshold-based alternation vs.\ classic 1:1
interleaving), hinge vs.\ BCE losses, and a weakened head-only discriminator —
with all variants reaching a $\sim$11\% relative gain by 800 steps. The best
configuration ($\lambda{=}0.3$, 1:1 alternation) keeps improving with longer
training and peaks at 7.4 by 5k steps — ahead of GRPO at the same budget —
but slowly gives gains back afterwards (6.9 at 50k), whereas GRPO climbs
monotonically to 7.7. We attribute the difference to anchoring: the adaptive
weight bounds the \emph{magnitude} of the adversarial force but provides no
positional anchor, so long-horizon training slowly erodes the peak, while
GRPO's KL trust region to the frozen stage-1 model lets improvements
accumulate stably.

\begin{table}[h]
\centering
\caption{Stage-2 objective ablation for the 8$\times$8 \method: teacher-forced
LAMBADA last-word exact-match (\%, 2{,}500 passages, 5 languages). All GAN
variants use the straight-through estimator through the discriminator and the
adaptive gradient-ratio weight; without it every $\lambda_m \in \{1,3,9\}$
diverges within 200 steps.}
\scalebox{0.9}{
\begin{tabular}{lrr}
\toprule
Stage-2 objective (113M, 8$\times$8) & steps & TF exact \\
\midrule
none (stage-1 baseline) & -- & 5.1 \\
GAN, adaptive $\lambda{=}0.1$, threshold alt. & 800 & 5.7 \\
GAN, adaptive $\lambda{=}0.3$, 1:1 alt. & 800 & 5.6 \\
GAN, adaptive $\lambda{=}0.3$, 1:1 alt. & 5000 & 7.4 \\
GAN, adaptive $\lambda{=}0.3$, 1:1 alt. & 50000 & 6.9 \\
GAN, adaptive $\lambda{=}0.3$, threshold alt. & 800 & 5.6 \\
GRPO & 5000 & 6.8 \\
GRPO & 50000 & \textbf{7.7} \\
\bottomrule
\end{tabular}
}

\label{stage2ablate}
\end{table}

\paragraph{Released GAN stage-2 model.} The best adversarial configuration
(adaptive $\lambda{=}0.3$, 1:1 alternation) is released as the official GAN
stage-2 checkpoint at its LAMBADA free-generation peak of 5{,}000 steps.
Full recipe: the generator and the context-aware patch discriminator are both
initialized from the 1M-step stage-1 backbone; the generator loss is
$\lambda_{\text{mle}}\,\mathcal{L}_{\text{rec}} + w\,\mathcal{L}_{\text{adv}}$ with
$\lambda_{\text{mle}}{=}9$ and the adversarial gradient rescaled each step to
$0.3\times$ the reconstruction gradient's norm at the generator's last layer;
BCE GAN loss with label smoothing $0.9$; strict 1:1 D/G alternation after a
50-step discriminator warm-up. Optimization: AdamW
($\beta{=}(0.9,0.95)$, weight decay $0.01$), constant learning rate $10^{-5}$
with 100 warm-up steps, batch size 128, bf16 mixed precision, 5{,}000 steps on
the stage-1 mC4 language mix (seed 42, 4$\times$H100).

\paragraph{Re-evaluation under the free-generation protocol.} The table above
uses the legacy teacher-forced exact-match metric, whose context windows leak
the first characters of the target word; under the paper's free-generation
OCR protocol (\cref{app:ocrjudge}) the picture is sharper. The GAN
recipe peaks at \textbf{6.12} LAMBADA accuracy at 5k steps and decays with
longer training (5.56 at 50k, measured with the generic transcription prompt),
essentially matching its own stage-1 backbone (6.00), while the released GRPO
stage-2 model reaches \textbf{10.40} under the identical protocol ($+70\%$).
On bAbI (dash-delimited 3-shot) the gap is the same shape: the GAN model scores
7.8, below the 11.4 of the stage-1 backbone it was trained from, whereas GRPO
reaches 15.6. The adversarial objective is thus
sufficient for local glyph realism -- enough for single-word answers -- but
provides no training signal on the model's own free-running trajectories,
which is precisely what the rollout GRPO stage supplies.

\section{Stage-2 free-generation RL details}
\label{app:stage2rldetails}

All three released stage-2 models (\cref{GLUE1Mtable,XNLI1M,babilamb}, rows
$stage2$) are trained with the rollout GRPO procedure of
\cref{sec:multilingual} on the stage-1 mC4 language mix, with unpacked
right-padded rendering (one document per sequence, \texttt{seq\_len} 180
blocks). The discriminator/generator gating uses an EMA (0.95) of the
discriminator's accuracy with thresholds $0.55/0.75$ and phase lengths bounded
to $[100, 400]$ steps. Common losses: PPO clip $0.2$, reconstruction anchor
$\lambda_{\text{rec}}{=}0.1$, group-relative advantages with std floor $10^{-2}$,
sampling temperature $1.0$, greedy-equivalent rollout length $K{=}3$ blocks
from a word onset with 48 context blocks, 8 rollout rows per micro-batch.
AdamW ($\beta{=}(0.9,0.95)$, wd $0.01$), constant LR with 100 warm-up steps.

\begin{table}[h]
\centering
\caption{Per-model rollout-GRPO hyperparameters of the released stage-2
models. The word gap is the all-white column run that marks a word boundary,
calibrated to each font size.}
\scalebox{0.9}{
\begin{tabular}{lccc}
\toprule
 & 113M (8$\times$8) & 116M (32$\times$32) & 477M (32$\times$32) \\
\midrule
initialization & GRPO 50k ckpt & 4-generation chain & 1-generation chain \\
batch / micro-batch & 128 / 32 & 32 / 8 & 128 / 8 \\
learning rate & $10^{-5}$ & $5\times10^{-6}$ & $10^{-5}$ \\
steps per generation & 2000 & 1500--4000 & 2000 \\
group size $G$ & 8 & 16 & 8 \\
reward blend $d_{\text{blend}}$ & 0.5 & 0.7 & 0.5 \\
policy weight $\lambda_{\text{pol}}$ & 26 & 10 & 10 \\
KL weight $\beta_{\text{KL}}$ & 0.02 & 0.05 & 0.05 \\
word-gap (px) & 3 & 12 & 12 \\
\bottomrule
\end{tabular}
}

\label{stage2rlhp}
\end{table}

\paragraph{Hyperparameter search.} The reported values were selected through
short pilot runs, scored by validation LAMBADA free-generation accuracy. The
search covered $d_{\text{blend}} \in [0, 0.7]$, $\lambda_{\text{tile}} \in
[0.3, 0.9]$, $\beta_{\text{KL}} \in [0, 0.5]$, $\lambda_{\text{pol}} \in [0,
100]$, $\lambda_{\text{rec}} \in [0, 1]$, group sizes $G$ from 2 to 256 and
learning rates from $2\times10^{-6}$ to $5\times10^{-5}$. Setting
$d_{\text{blend}} = 0$ or $\lambda_{\text{rec}} = 0$ removes the discriminator
or the reconstruction anchor entirely, so both ablations are contained in the
search. Performance improved with $\beta_{\text{KL}}$ up to the reported
values.

\paragraph{Iterated warm chains.} The 32$\times$32 models are the product of
\emph{iterated} RL: each generation re-initializes both players from the
previous generation's champion checkpoint and re-runs the same recipe with a
fresh data-order seed. Roughly one seed in five improves on its parent, but
the expectation rises generation over generation; the 116M model used five
generations, after which a sixth showed uniform regression (over-chaining
erosion). The 477M model needed a single extra generation on top of its
first-stage champion.

\paragraph{Weight soups.} Each released checkpoint is a uniform or
lightly-tilted average of champion checkpoints: the 477M release averages two
checkpoints of one run ($0.6/0.4$), the 116M 32$\times$32 release averages the
two strongest sibling checkpoints of the final generation ($0.5/0.5$), and the
113M release averages the champion run's three peak checkpoints (equal
weights). Same-run or sibling averages at these ratios consistently helped;
averaging three or more distinct runs, or runs from different recipe families,
consistently diluted.

\paragraph{Reward design must match patch geometry.} Extending the stage-2
search to the packed-retrained backbones shows that the effective reward
\emph{type} depends on patch resolution. On 8$\times$8 patches the
discriminator reward remains the best choice (as above); on 32$\times$32
patches every discriminator-based variant we tried degrades the model, while a
\emph{ground-truth blended reward} --- $0.7\times$ (fraction of full-height,
8-px-wide column tiles predicted entirely correctly against the real next
patch) $+\,0.3\times$ (per-pixel match fraction), with no discriminator at all
--- yields large, reproducible gains. We attribute the pairing to patch
information density: given the context, the next 32$\times$32 patch is nearly
deterministic, so an oracle reward is informative whereas real/fake
discrimination is trivially easy and its gradient direction is uncorrelated
with correctness; on 8$\times$8 patches the converse holds. Within the blend,
the tile term supplies an all-or-nothing signal aligned with exact-match and
the pixel term preserves calibration: with the tile term alone, gains on some
languages came at the expense of others (Spanish regressed by 21\%), while the
blend improves every language and also raises last-word pixel accuracy.
The \methodnospace$_{\text{stage2}}$ rows of \cref{babilamb} report the resulting
models; both 32$\times$32 backbones peak within 350--700 policy steps, and each
recipe was replicated across three to six runs (spread $\pm$0.7 points). All
columns follow the free-generation OCR protocol of \cref{app:ocrjudge}.

\section{Text recognition}
\label{app:ocrjudge}
Evaluating free generation requires reading text out of the generated patches.
\citet{tai2024pixar} used classical OCR engines (PaddleOCR and Tesseract) for
this purpose, but these engines are unreliable on low-resolution binary
renderings and are only available for a subset of our languages, so we replace
them with a single multilingual vision-language judge, Qwen3.6-35B-A3B, applied
uniformly to every model and language. The protocol has four parts.

\paragraph{Pixel-exact shortcut.} Before any judging, we compare the generated
patches against the rendering of the gold continuation. If the generated pixels
match the gold rendering exactly, up to the word boundary, the sample is scored
correct without invoking the judge. This shortcut is judge-independent and
accounts for roughly a quarter of the correct answers of the 477M model; the
remaining samples are passed to the judge, so judging errors can only cost
accuracy, never manufacture it.

\paragraph{Image preparation.} The patches to be read are concatenated into a
strip, upscaled four times with nearest-neighbour interpolation, and pasted at
the centre of a square white canvas, following the observation of
\citet{tai2024pixar} that bare thin strips degrade recognition. The judge is
then asked to transcribe the image, with a prompt stating that the text is
pixel-rendered and may be in any of several languages. We verified that this
language-aware prompt changes accuracy by at most $0.6$ points on either
benchmark relative to a generic transcription prompt, for every model and
training stage.

\paragraph{What the judge sees.} On bAbI the judge reads the generated region
only. On LAMBADA it reads the generation preceded by a $16$-block anchoring
context, and the transcription is cut at the last prompt fragment the judge
reproduced, so context words cannot be counted as an answer. This matters:
if the judge is instead shown the entire prompt image, it can infer the answer
from the visible question rather than transcribe what the model generated,
which inflates the scores of weak models to the point of reversing the ordering
between our stage-1 and stage-2 models on bAbI. Restricting the judge's view to
the generated region removes this leakage, and the resulting ordering agrees
with the judge-independent pixel-exact metric for every model.

\paragraph{Matching.} Transcriptions are lowercased and whitespace-normalised,
and glyph pairs that our renderer makes pixel-identical after descender
clipping (\texttt{q}/\texttt{g}, \texttt{0}/\texttt{o}, \texttt{1}/\texttt{l})
are folded together before comparison. A bAbI answer counts as correct if the
gold word appears among the first three transcribed words, which absorbs stray
punctuation and digits at the start of a noisy transcription; LAMBADA requires
the gold last word.

\section{Orthographic attacks}
\begin{figure}[!t]
    \centering
    \includegraphics[width=8cm]{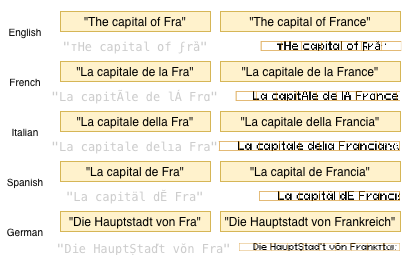}
    \caption{Examples of 5 latin languages contains in pretraining dataset. 20\% letters in these letters are changed by confused one.}
    \label{confuse_mixar}
\end{figure}

\cref{confuse} reports the accuracy of every model at each attack ratio, and \cref{confuse_mixar} shows how the attack looks for the other alphabetic languages of our pretraining mix. The confused letters in this image are generated through the \href{https://pypi.org/project/confusables/}{confusables package} in python. 20\% letters in these sentenses are changed into the confused ones. 

\section{0.1M steps results of GLUE}

Table \ref{GLUE100ktable} shows the performance of \pixar and \method trained on 0.1M steps and finetuned on GLUE tasks.

\section{Pretraining dataset information}
Table \ref{tab:dataset_stats} shows the number of patches we used for the pretraining stage 1.

\begin{table*}[htbp]
\centering
\caption{This table shows the hyperparameters applied in the GLUE benchmark.} 
\scalebox{0.8}{
\begin{tabular}{c|ccccccccc}
\hline
stage1 & MNLI &  QQP & QNLI & SST-2 & COLA & STSB & MRPC & RTE & WNLI \\
\hline
85M \pixarnospace$_{\text{stage1}}$ lr & 3e-5 & 3e-5 &3e-5 &3e-5 &3e-5 &3e-5 &6e-5 &3e-5 &3e-5 \\
116M \methodnospace$_{\text{stage1}}$ lr & 3e-5 & 3e-5 &3e-5 &3e-5 &3e-5 &3e-5 &6e-5 &3e-5 &6e-5 \\
477M \methodnospace$_{\text{stage1}}$ lr & 3e-5 & 3e-5 &3e-5 &3e-5 &6e-5 &3e-5 &6e-5 &3e-5 &6e-5 \\
Weight decay & 0.1 & 0.1 & 0.1 & 0.01 & 0.01 & 0.01 & 0.01 & 0.01 & 0.01 \\
Optimizer & \multicolumn{9}{c}{AdamW}\\
Warmup & \multicolumn{9}{c}{Linear warmup} \\
Warmup steps & 1000 & 1000 & 500 & 200 & 50 & 100 & 20 & 50 & 2 \\
$\beta_1$ & \multicolumn{9}{c}{0.9}\\
$\beta_2$ & \multicolumn{9}{c}{0.95}\\
Random seed & \multicolumn{9}{c}{42} \\
Batch size & 256 & 256 & 256 & 256 & 256 & 32 & 64 & 32 & 128 \\
Max steps & 8000 & 8000 & 4000 & 2000 & 500 & 2000 & 500 & 500 & 20\\
evaluation freq. & 500 & 500 & 200 & 200 & 100 & 100 & 50 & 50 & $\sim$1 epoch\\

\hline
stage2 & MNLI &  QQP & QNLI & SST-2 & COLA & STSB & MRPC & RTE & WNLI \\
\hline
85M \pixarnospace$_{\text{stage2}}$ lr & 3e-5 & 3e-5 &3e-5 &3e-5 &3e-5 &3e-5 &3e-5 &6e-5 &3e-5 \\
116M \methodnospace$_{\text{stage2}}$ lr & 3e-5 & 3e-5 &3e-5 &3e-5 &6e-5 &3e-5 &6e-5 &3e-5 &6e-5 \\
477M \methodnospace$_{\text{stage2}}$ lr & 3e-5 & 3e-5 &3e-5 &3e-5 &1e-4 &3e-5 &8e-5 &9e-5 &3e-6 \\
Weight decay & 0.1 & 0.1 & 0.1 & 0.01 & 0.01 & 0.01 & 0.01 & 0.01 & 0.01 \\
Optimizer & \multicolumn{9}{c}{AdamW}\\
Warmup & \multicolumn{9}{c}{Linear warmup} \\
Warmup steps & 1000 & 1000 & 500 & 200 & 50 & 100 & 20 & 50 & 2 \\
$\beta_1$ & \multicolumn{9}{c}{0.9}\\
$\beta_2$ & \multicolumn{9}{c}{0.95}\\
Random seed & \multicolumn{9}{c}{42} \\
Batch size & 256 & 256 & 256 & 256 & 256 & 32 & 64 & 32 & 128 \\
Max steps & 8000 & 8000 & 4000 & 2000 & 500 & 2000 & 500 & 500 & 20\\
evaluation freq. & 500 & 500 & 200 & 200 & 100 & 100 & 50 & 50 & $\sim$1 epoch\\
\hline
\end{tabular}
}

\label{HyperparameterGLUE}  
\vspace{20pt}
\end{table*}

\begin{table*}[htbp]
\centering
\caption{This table shows the hyperparameters used in XNLI downstream tasks.} 
\scalebox{0.6}{
\begin{tabular}{cccccccccccc}
\hline
Models & lr & Weight decay & Optimizer & Warmup & Warmup steps & $\beta_1$  & $\beta_2$  & Random seed & Batch size & Max steps & evaluation freq.\\

\hline
\methodnospace$_{\text{stage1}}$ & 3e-5 &  0.1 & AdamW &  Linear warmup & 1000 & 0.9 & 0.95 & 42  & 256 & 8000 & 500 \\
\methodnospace$_{\text{stage2}}$  & 3e-5 &  0.1 & AdamW &  Linear warmup & 1000 & 0.9 & 0.95 & 42  & 256 & 8000 & 500  \\
\hline
\end{tabular}
}

\label{HyperparameterXNLI}  
\vspace{20pt}
\end{table*}

\begin{table*}[htbp]
\centering
\caption{This table shows the hyperparameters used in SIB downstream tasks.} 
\scalebox{0.6}{
\begin{tabular}{cccccccccccc}
\hline
Models & lr & Weight decay & Optimizer & Warmup & Warmup steps & $\beta_1$  & $\beta_2$  & Random seed & Batch size & Max steps & evaluation freq.\\

\hline
\methodnospace$_{\text{stage1}}$ & 5e-5 &  0.01 & AdamW &  Linear warmup & 50 & 0.9 & 0.95 & 42  & 128 & 500 & 25 \\
\hline
\end{tabular}
}

\label{HyperparameterSIB}  
\end{table*}

\begin{table*}[htbp]
\centering
\caption{Accuracy under the orthographic attack for each attack ratio (English LAMBADA and bAbI, free-generation protocol, lookalike glyphs restricted to the bundled fonts). The \method entries are the stage-2 models; MGPT-2, which shares our pretraining corpus, receives the identical attacked text. Prior-work numbers (e.g., \pixar) are not comparable under this protocol and are omitted.}
\scalebox{0.8}{
\begin{tabular}{ccccccccc}
\hline
\textbf{attack} & \multicolumn{4}{c}{\textbf{LAMBADA}} & \multicolumn{4}{c}{\textbf{bAbI}} \\
\cline{2-9}
\textbf{ratio}  & \textbf{GPT-2} & \textbf{MGPT-2} & \textbf{\methodnospace} (116M) & \textbf{\methodnospace} (477M) & \textbf{GPT-2} & \textbf{MGPT-2} & \textbf{\methodnospace} (116M) & \textbf{\methodnospace} (477M)  \\
\hline
0.0  & 32.6 & 26.2 & 9.2 & 13.4 & 41.4 & 38.0 & 16.6 & 34.6 \\
0.01 & 30.0 & 24.6 & 9.0 & 12.6 & 43.2 & 36.2 & 17.0 & 34.6 \\
0.05 & 19.0 & 15.6 & 8.8 & 13.2 & 35.8 & 24.2 & 15.0 & 23.8 \\
0.1  &  8.4 &  7.2 & 7.6 &  9.6 & 23.0 & 14.0 & 13.2 & 18.6 \\
0.2  &  1.6 &  2.2 & 4.8 &  7.4 &  8.0 &  4.2 & 11.2 & 11.4 \\
0.3  &  1.0 &  1.0 & 4.0 &  4.4 &  1.0 &  2.0 &  8.4 &  8.0 \\
0.4  &  0.0 &  0.0 & 3.4 &  2.6 &  0.0 &  0.8 &  6.4 &  3.6 \\
0.5  &  0.0 &  0.2 & 1.8 &  1.8 &  0.0 &  0.2 &  4.4 &  2.2 \\
\hline
\end{tabular}
}

\label{confuse}
\end{table*}

\begin{table*}[htbp]
\centering
\caption{This table shows the comparison of \pixar and \method with the patch size of 8$\times$8 and 32$\times$32 trained on 0.1M steps and finetuned on GLUE tasks. Matthew's correlation is used for COLA and Spearman's $\rho$ is applied for STSB. The F1 score is for MRPC and QQP. Other tasks used accuracy.} 
\setlength{\tabcolsep}{2pt}
\scalebox{0.8}{
\begin{tabular}{ccccccccccccc}
\hline
\multirow{2}*{Models} & \multirow{2}*{Parameters} & Patch size& MNLI-m/mm &  QQP & QNLI & SST-2 & COLA & STSB & MRPC & RTE & WNLI & \multirow{2}*{AVG}\\
\cline{4-12}
 &  &  & 392k  &  363k & 108k & 67k  & 8.5k & 5.7k & 3.5k & 2.5k & 635 &  \\
\hline 
PIXAR$_{\text{stage1}}$ & 85M & 8$\times$8 &  76.4 / 77.6 &  84.3 & 84.2 &  88.0 & 27.7 & 81.2 & 82.5 & 58.5 & 56.3 & 71.6 \\
\methodnospace$_{\text{stage1}}$ & 116M & 32$\times$32 & 72.5 / 72.6 & 83.2 & 82.9 & 83.6 & 17.6 & 80.6 & 83.1 & 61.0 & 56.3 & 69.3 \\
\methodnospace$_{\text{stage1}}$ & 474M & 8$\times$8 & 78.6 / 78.0 & 85.6 & 85.1 & 88.8 & 28.5 & 82.9 & 82.9 & 65.3 & 56.3 & 73.2 \\
\methodnospace$_{\text{stage1}}$ & 477M & 32$\times$32 & 76.1 / 76.1 & 85.3 & 84.9 & 87.7 & 23.2 & 82.8 & 83.2 & 63.9 & 56.3 &  72.0\\

\hline
\end{tabular}
}

\label{GLUE100ktable}  
\end{table*}

\begin{table*}[htbp]
\centering
\caption{Patch-level data statistics for the \method pretraining languages in the multilingual mC4 dataset, including average characters and GPT-2 tokens per patch.} 
\scalebox{1}{
\begin{tabular}{cccc}
\hline
Dataset & Number of patches &chars/patch& GPT2 tok/patch \\
\hline
Chinese mc4 & 3,345,800,998 & 2.47 & 4.40\\
English mc4 & 17,387,543,604 & 3.26 & 0.81 \\
French mc4 & 23,633,038,801 & 3.28 & 1.16 \\
German mc4 & 28,550,835,190 & 3.20 & 1.19 \\
Italian mc4 & 12,161,721,091 & 3.35 & 1.20 \\
Spanish mc4 & 33,542,982,712 & 3.25 & 1.16 \\
Japanese mc4 & 17,515,002,689 & 2.68 & 3.67 \\
Korean mc4 & 2,103,074,913 & 2.69 & 4.94 \\
\hline
\end{tabular}
}

\label{tab:dataset_stats}  
\end{table*}

\clearpage

\begin{figure*}[htp]
    \centering
    \includegraphics[width=15cm]{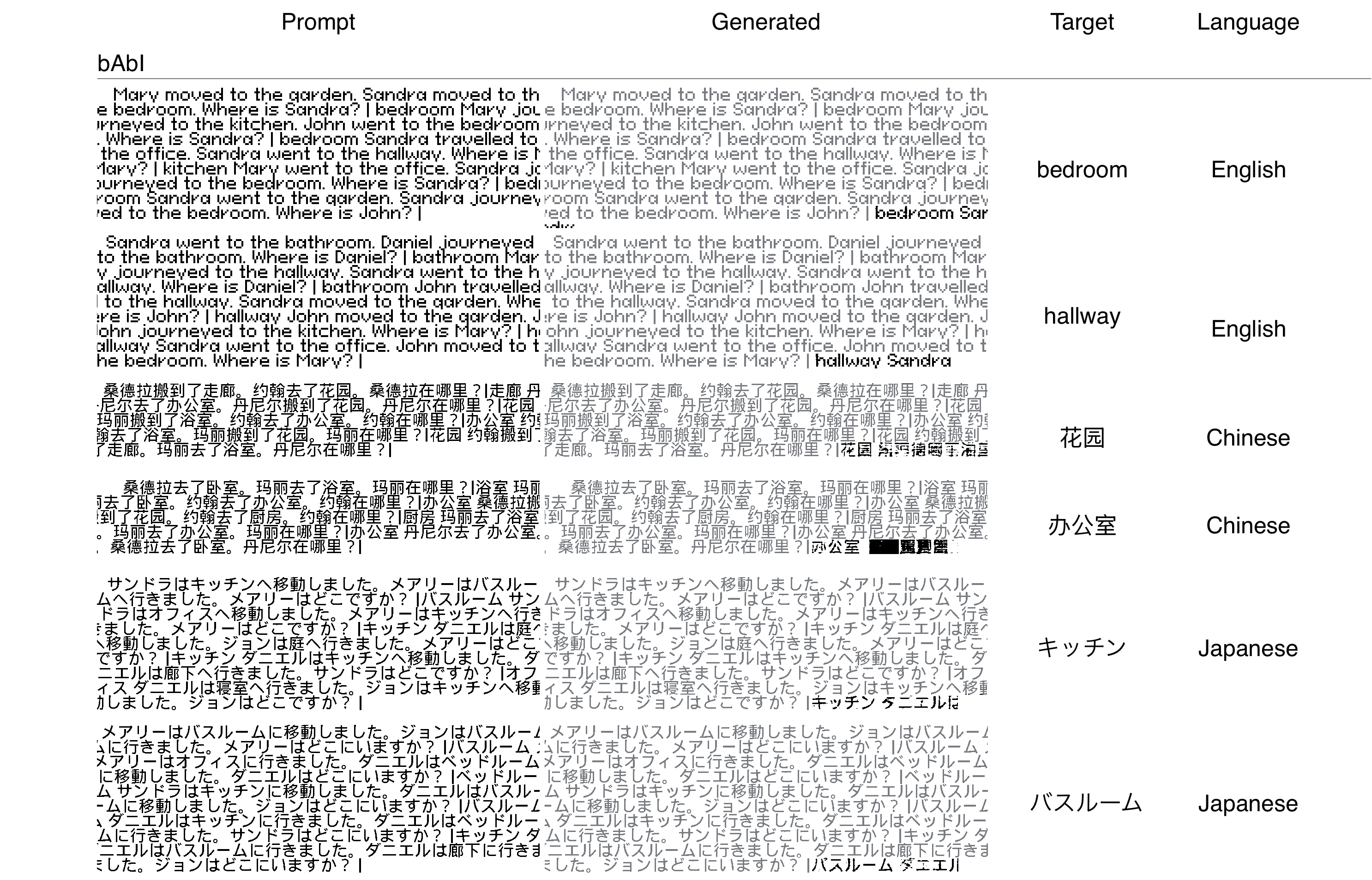}
    \caption{This image shows the correct samples of English, Chinese and Japanese bAbI tasks}
    \label{babi116good}
\end{figure*}

\begin{figure*}[htp]
    \centering
    \includegraphics[width=15cm]{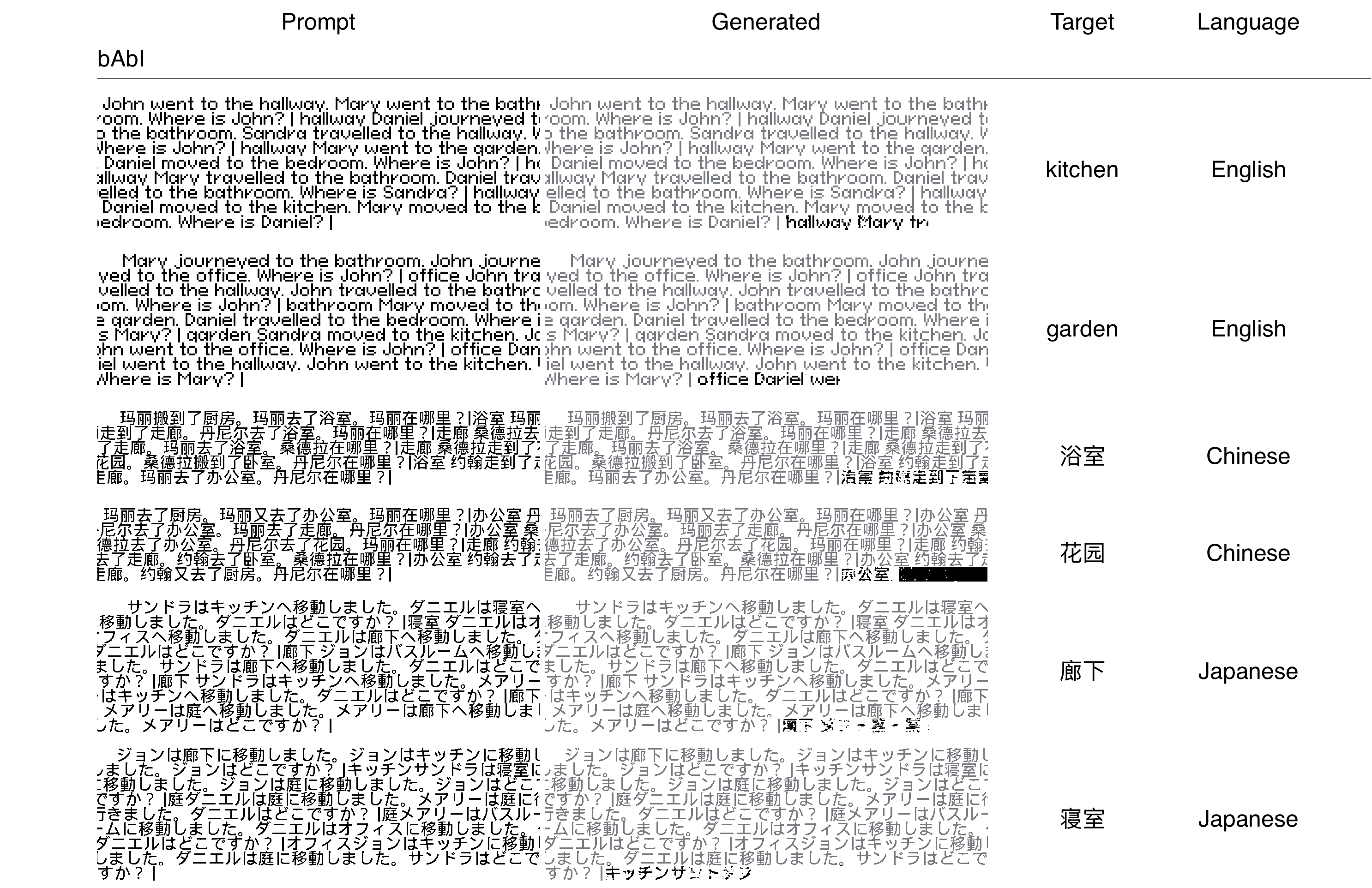}
    \caption{This image shows the wrong samples of English, Chinese and Japanese bAbI tasks}
    \label{babi116bad}
\end{figure*}

\begin{figure*}[htp]
    \centering
    \includegraphics[width=15cm]{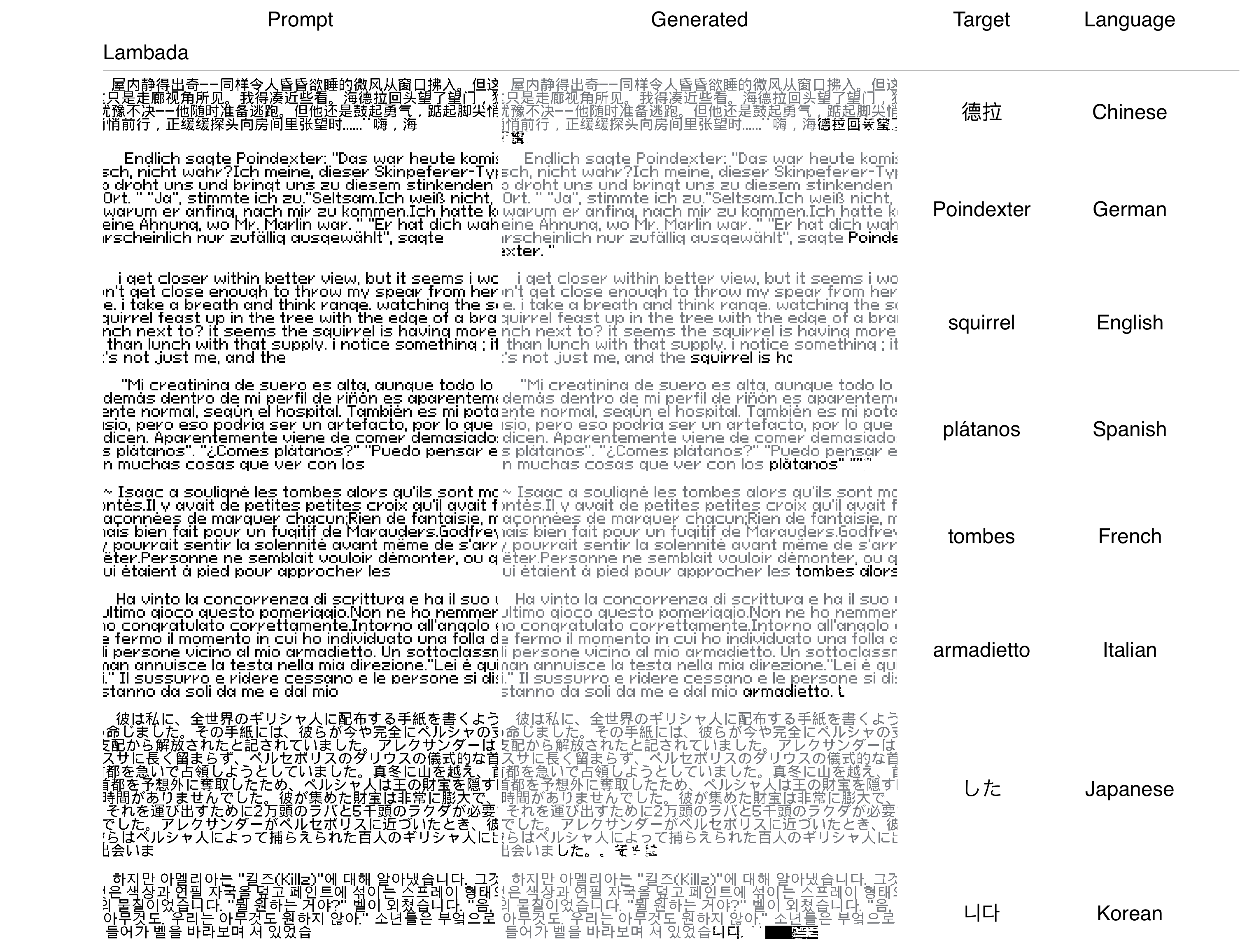}
    \caption{This image shows some correct samples of LAMBADA task of eight pretraining languages}
    \label{lamb116good}
\end{figure*}

\begin{figure*}[htp]
    \centering
    \includegraphics[width=15cm]{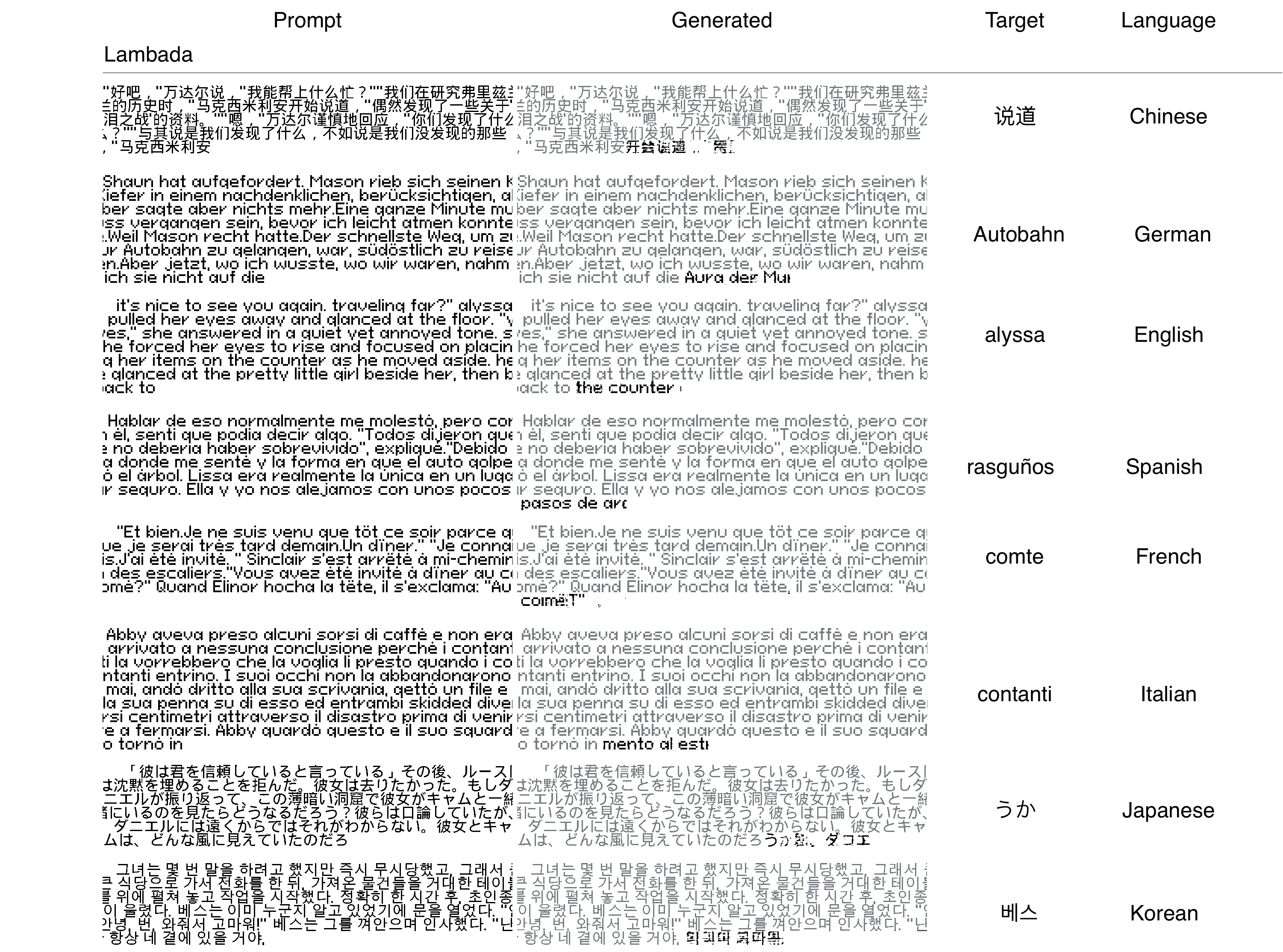}
    \caption{This image shows some wrong samples of LAMBADA task of eight pretraining languages}
    \label{lamb116bad}
\end{figure*}

\end{document}